\pdfoutput=1

\documentclass[11pt]{article}

\usepackage{acl}

\usepackage{times}
\usepackage{latexsym}
\usepackage{amsmath}
\usepackage{booktabs}

\usepackage[T1]{fontenc}

\usepackage[utf8]{inputenc}

\usepackage{microtype}

\usepackage{inconsolata}

\usepackage{graphicx}

\title{SpeechIQ: Speech-Agentic Intelligence Quotient Across Cognitive Levels in Voice Understanding by Large Language Models}

\author{
Zhen Wan$^1$  \hspace{1em}
Chao-Han Huck Yang$^2$ \hspace{1em} 
Yahan Yu$^1$ \hspace{1em}
{\bf Jinchuan Tian$^3$} \hspace{1em} \\
{\bf Sheng Li$^4$ } \hspace{1em}
{\bf Ke Hu$^2$ } \hspace{1em} 
{\bf Zhehuai Chen$^2$ } \hspace{1em} 
{\bf Shinji Watanabe$^3$ } \hspace{1em} 
{\bf Fei Cheng$^1$ }\hspace{1em} \\
{\bf Chenhui Chu$^1$ }\hspace{1em}
{\bf Sadao Kurohashi$^1$ }\hspace{1em}\\
$^1$Kyoto University \hspace{1em}
$^2$NVIDIA \hspace{1em}
\\
$^3$Carnegie Mellon University  \hspace{1em}
$^4$Institute of Science Tokyo \hspace{1em}
\\
\texttt{zhenwan@nlp.ist.i.kyoto-u.ac.jp} \quad
\texttt{hucky@nvidia.com} \\
}

\begin{document}
\maketitle
\begin{abstract}
We introduce Speech Intelligence Quotient (SpeechIQ) as a new form of human cognition-inspired evaluation pipeline for voice understanding large language models (LLM\textsubscript{Voice}), designed to assess their voice understanding ability. 
Moving beyond popular voice understanding metrics such as word error rate (WER), SpeechIQ examines LLM\textsubscript{Voice} across three cognitive levels motivated by Bloom’s Taxonomy: (1) Remembering (i.e., WER for verbatim accuracy); (2) Understanding (i.e., similarity of LLM's interpretations); and (3) Application (i.e., QA accuracy for simulating downstream tasks). We demonstrate that SpeechIQ not only quantifies voice understanding abilities but also provides unified comparisons between cascaded methods (e.g., ASR-LLM) and end-to-end models, identifies annotation errors in existing benchmarks, and detects hallucinations in LLM\textsubscript{Voice}. Our framework represents a first-of-its-kind intelligence examination that bridges cognitive principles with voice-oriented benchmarks, while exposing overlooked challenges in multi-modal training. Our Speech-IQ leaderboard is hosted at \url{huggingface.co/spaces/nvidia/Speech-IQ-leaderboard}.

\end{abstract}

\section{Introduction}
The rapid rise of voice understanding Large Language Models (LLM$_\text{Voice}$) that can process voice input via the speech portal has ushered in a new paradigm of human-machine interaction~\cite{reddy1988foundations}, where LLM$_\text{Voice}$ process spoken instructions, infer semantic intent, and execute downstream tasks~\cite{jelinek1991dynamic, jurafsky1995using,zue2000conversational,tur2005combining, mesnil2014using, kawahara2019spoken, huang2024dynamic, Mahmood_2025}. LLM$_\text{Voice}$ bridge \textit{speech} and \textit{language intelligence}~\cite{sparks1996role}, enabling applications from voice assistants to interactive robots. A fundamental prerequisite for deploying robust LLM$_\text{Voice}$ lies in establishing reliable evaluation metrics for voice understanding. These metrics ensure that the system accurately comprehends voice inputs.
Automatic speech recognition (ASR) is the central component of existing cascaded\footnote{We refer to two-pass based neural ASR-LM systems~\cite{hoffmeister2016language, sainath19_interspeech, huang20f_interspeech}, which have been widely adopted in the current voice assistants, including Siri, Alexa, and Google Assistant.} LLM$_\text{Voice}$ systems, transcribing voice into text for subsequent LLM processing. Consequently, the research community primarily assesses voice understanding using transcription error metrics~\cite{hunt1990figures, klakow2002testing}, with Word Error Rate (WER) standing as the de facto standard.

However, WER primarily measures the lexical recall of ASR, its transcribed quality does \textbf{not} fully capture the semantic understanding~\cite{he2011word, chen2018spoken, szymanski2020we} and task completion~\cite{van2020assessing} capabilities of LLMs via their speech portal. For instance, WER treats all lexical errors equally, ignoring how transcription inaccuracies propagate to higher-level comprehension (\textit{i.e.}, semantic understanding) and executions (\textit{i.e.}, downstream tasks). 

\begin{figure}[t!]
  \includegraphics[width=\linewidth]{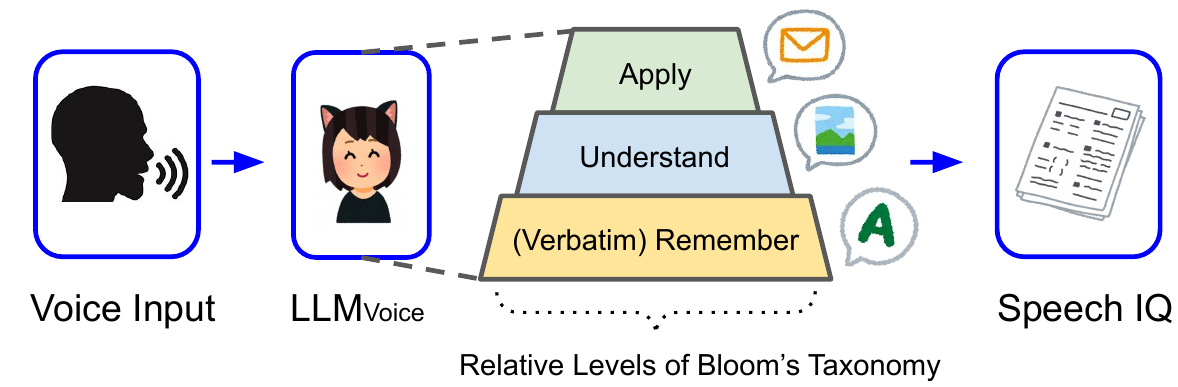}
  \caption{An overview of cognitive levels in voice understanding large language model systems related to the first three foundational hierarchies of Bloom's Taxonomy. We design a corresponding examination to measure SpeechIQ as a detailed pipeline in Figure~\ref{result: method}.}
  \label{result: overview}
\end{figure}


Meanwhile, a growing trend in LLM$_\text{Voice}$ research is the shift toward end-to-end multi-modal models~\cite{rubenstein2023audiopalm, radhakrishnan2023whispering, maiti2024voxtlm, hu2024wavllmrobustadaptivespeech, defossez2024moshi, chen2024salm}, which bypass explicit ASR transcription and directly map audio inputs to task-specific outputs (e.g., answers, actions). While this paradigm simplifies pipelines and reduces error cascades, its end-to-end voice service obviates the need for generating an intermediate transcript, rendering traditional WER-based evaluation obsolete~\cite{kuo20_interspeech, tuske21_interspeech}. This creates a critical methodological gap: without a reliable and potentially unified metric, comparing voice understanding across architectural choices, such as the modularity of ``\textit{cascaded neural systems}'' and the efficiency of ``\textit{end-to-end mutlimodal models},'' remains insufficiently characterized.



The above two limitations mirror a fundamental challenge in assessing artificial intelligence: as Figure~\ref{result: overview}, human-like intelligence is inherently hierarchical. Drawing on Bloom’s Taxonomy~\cite{bloom1969taxonomy,10.1145/1553374.1553380}, a cognitive framework where skills ascend from basic remembering to advanced creating, we posit that LLM$_\text{Voice}$ should be evaluated through analogous levels. In this framework, WER represents the lowest level by measuring verbatim recall while neglecting higher-order abilities such as semantic understanding and task-solving. For instance, two transcripts with identical WER may differ drastically in meaning~\cite{szymanski2023aren}, leading to divergent LLM$_\text{Voice}$ responses or failed instructions.

To address this gap, we propose a hierarchical IQ test for LLM$_\text{Voice}$ that evaluates intelligence across three fundamental levels (refer to limitations) aligned with Bloom’s Cognitive Taxonomy: (1) Remember in verbatim: Lexical accuracy is quantified using WER; 
(2) Understand: Semantic consistency is assessed by comparing LLM responses to ASR outputs and ground-truth transcripts. LLMs are prompted to infer the background context (\textit{e.g.}, domain, speaker intent) and generate a summary of the speech, then measure hidden states similarity between ASR-derived and ground truth-derived responses;
(3) Apply: Task-solving capability is tested via multi-choice question-answering (QA) pairs constructed from ground-truth transcripts. LLM$_\text{Voice}$ answers questions based on speech inputs, with accuracy reflecting real-world utility.
Then, we draw inspiration from Raven’s Progressive Matrices~\cite{raven1998raven, John2003} (\textit{i.e.}, a human IQ test) that aggregates performance across various dimensions into a single score. 

Beyond overall evaluation, our SpeechIQ framework yields two key insights:

\begin{itemize}
    \item Cascade systems (ASR+LLM) outperform end-to-end models under similar scaling, revealing modality interference in joint speech-text training.
    \item Cross-model QA consistency exposes ground-truth errors. By isolating questions most LLM$_\text{Voice}$, cannot solve, we create an ``unanswerable'' set that helps detect hallucinations~\cite{frieske2024hallucinationsneuralautomaticspeech,Koenecke2024CarelessWS} and refine benchmark annotations.
\end{itemize}

Our work advances LLM$_\text{Voice}$ evaluation by merging cognitive principles with practical metrics, revealing hidden challenges in multimodal inference (\textit{e.g.}, hallucination inheritance), and offering tools to build more robust systems. Code and data will be publicly available.
\section{Related Work}
\subsection{ASR Evaluation}
ASR systems have long been evaluated using lexical similarity metrics such as WER, which quantifies the Levenshtein distance~\cite{10.1145/375360.375365} between ASR outputs and reference transcripts. Metrics like character error rate~\cite{mackenzie2002character} (CER), sentence error rate~\cite{juffs1996garden} (SER), and translation error rate (TER) extend this framework to the sentence or translation level but share WER’s core limitation: treating all errors equally, regardless of semantic impact. While effective for benchmarking transcription fidelity, these metrics do not fully capture how errors propagate to downstream tasks.


Recent work mitigates WER’s semantic insensitivity by incorporating sentence embeddings into evaluation. Sentence similarity-based metrics~\cite{kim2021semanticdistancenewmetric,kim2022evaluatinguserperceptionspeech} and BERTScore~\cite{zhang2020bertscoreevaluatingtextgeneration} leverage pre-trained language models to compute semantic correspondence between ASR hypotheses and ground-truth references. Others propose task-specific metrics, such as Medconcept WER~\cite{adedeji2024soundhealthcareimprovingmedical} which arranges more weights on \textit{medical entities} in WER computation or severity score~\cite{whetten-kennington-2023-evaluating} that involves sentiment similarity in the evaluation. However, these methods are either confined to specific tasks or require \textbf{curated labeled data}, limiting their generalizability to open-domain LLM$_\text{Voice}$.

Efforts to develop hybrid metrics that combine error rate and semantic similarity, such as H\_eval~\cite{sasindran2023hevalnewhybridevaluation}, and Sema~\cite{Sasindran_2024}, further enrich the evaluation landscape. Concurrently, machine translation inspired metrics like BLEU Score~\cite{papineni-etal-2002-bleu} and COMET~\cite{rei-etal-2020-comet}, have started to be adapted for ASR to assess fluency and pragmatic adequacy. However, these approaches still emphasize text-to-text alignment between ASR outputs and references, rather than from a view of influencing the responses and actions of downstream LLMs, motivating the proposed SpeechIQ as a multi-dimensional metric.

\subsection{LLM$_\text{Voice}$ Understanding Systems}
In general, speech-based LLMs encompass both the understanding of voice signals and the generation~\cite{agostinelli2023musiclm, borsos2023audiolm, pmlr-v235-yang24x} of vocal or general audio outputs. In this work, we focus exclusively on \textbf{voice understanding tasks} with text outputs as one first step to examine voice intelligence~\cite{sparks1996role} or its potential voice world model~\cite{ha2018recurrent, matsuo2022deep} via LLM backbone, which we denote as LLM$_\text{Voice}$. LLM$_\text{Voice}$ understanding systems are designed to interact~\cite{shah2018building, lopez2018alexa} with users and execute task-oriented instructions via speech-based inputs. This form of ``voice-to-text'' architecture can be broadly categorized into three main types, all of which achieve state-of-the-art ASR quality on public leaderboards~\cite{radford2022robustspeechrecognitionlargescale, Chen2023HyPoradiseAO, puvvada2024moreaccuratespeechrecognition}:

\paragraph{(1) Cascaded ASR + LLM}
This approach follows a traditional pipeline-based structure, where an ASR model~\cite{watanabe2018espnetendtoendspeechprocessing, gulati20_interspeech, radford2022robustspeechrecognitionlargescale} transcribes the spoken input, and the 1-best transcription is passed to textual LLM as one operation system~\cite{wu2024just,dighe2024leveraging} for direct response generation.

\begin{figure*}[t]
  \includegraphics[width=1.0\linewidth]{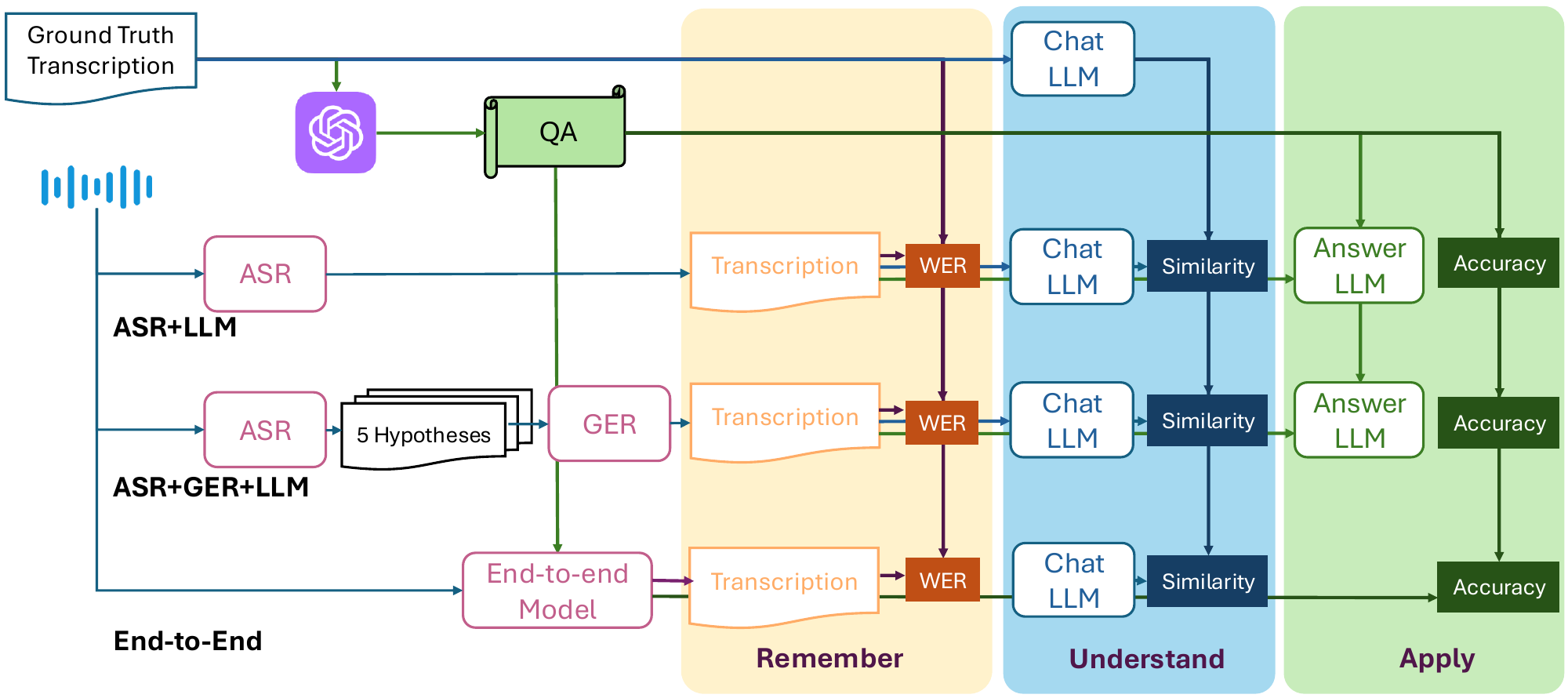} \hfill
  \caption {\textbf{Overview of Speech IQ (SIQ) test}. We compared cognition-inspired categories of LLM$_\text{Voice}$ represented in three rows, with the right columns denoting the SIQ sub-tests. The truth reference will be used in the sub-tests.}
  \label{result: method}
\end{figure*}

\paragraph{(2) Cascaded ASR Hypotheses + Generation Error Correction (GER) + LLM}
In this approach, the ASR model first generates multiple possible transcriptions (\textit{i.e.}, hypotheses) for a given speech input, where ``texts'' are then treated as input features for a many-to-one GER post-editing module~\cite{Yang_2023, velikovich2024spelling, hori2025delayed}. The GER module refines these hypotheses to determine the most accurate  transcription, which is subsequently processed by the LLM.

\paragraph{(3) End-to-End Multi-Modal Models}
These models are inherently designed to support both speech and text modalities~\cite{rubenstein2023audiopalm, hu2024wavllmrobustadaptivespeech, lyu2023macawllmmultimodallanguagemodeling, zhang2024speechgptgenscalingchainofinformationspeech}. Unlike cascaded approaches, they process audio inputs directly and generate textual outputs \textbf{without } requiring intermediate ASR transcriptions. This paradigm leverages a multimodal learning framework (\textit{i.e.}, a speech embedding encoder~\cite{radhakrishnan2023whispering} or tokenizer~\cite{borsos2023soundstorm, zhang2024speechtokenizer}) to achieve end-to-end speech understanding and reasoning. 

In summary, we carefully selected and examined aforementioned three ASR-LLM systems. System 1 serves as the most cost-effective solution~\cite{196553} for voice interactions with LLMs. System 2 is designed to decode texts that offer richer information~\cite{chan-etal-2023-ic3,yang2024large} for textual applications. System 3 highlights a low-latency, multimodal embedding injection approach for large language models~\cite{openai2024gpt4ocard, geminiteam2024gemini15unlockingmultimodal}. Next, we introduce our proposed method, which corresponds to Bloom's cognitive taxonomy.

\section{SpeechIQ: Speech Intelligence Quotient}



We will subsequently introduce each level of the test, from the remember level to both the understand level and apply level. At the end of this section, we will discuss the computation of the final SpeechIQ. An overview of our workflow is in Figure~\ref{result: method}.
\subsection{Remember: WER Metric}
The remember level assesses the ability to reproduce spoken content accurately. WER serves as a natural metric since it measures the verbatim discrepancy between ASR output and the ground truth at a granular level, with the advantage that it captures even the smallest differences, making it highly sensitive to transcription errors. 



\subsection{Understand: Similarity Metric}
The understand level assesses whether the ASR-generated transcript effectively conveys the intended meaning of the original speech. This is particularly important when ASR output serves as input to the LLM, and a minor transcription difference could lead to significant semantic variations to LLMs.
For instance, two transcriptions with identical WER scores can have vastly different meanings:
\begin{itemize} 
\item Ground truth: "I feel pain in the lower back." 
\item ASR 1: "I feel \textcolor{red}{like} pain in the \textcolor{red}{\_} back." 
\item ASR 2: "I feel \textcolor{red}{painting} in the \textcolor{red}{world} back." 
\end{itemize}
Although both have the same WER = 29\%, they convey different meanings and lead to drastically different LLM responses.

To quantify the impact of ASR errors on semantic comprehension, we measure the deviation in LLM responses caused by ASR transcription errors. Specifically, inspired by recent work \citep{jiang-etal-2024-scaling,inproceedings} showing that instructing LLMs to generate\textit{ one word} is a simple and straightforward approach to representing semantic meanings, we evaluate LLM-generated responses via two key questions:
(1) $b$: The background scenario of the speech [] in \textit{one word} is.
This helps resolve ambiguities and measure whether the ASR content preserves core contextual meaning;
(2) $s$: The summary of this speech [] in \textit{one word} is.
This assesses the extent to which ASR errors affect the LLM’s high-level understanding. 
Instead of directly comparing LLM-generated words, we use the last layer of hidden states of LLM for generating the next token as embeddings and compute their cosine similarity with the embeddings generated from the ground truth transcription:

\begin{equation} \text{Sim}\_b = \cos(\mathcal{M}\_b(\text{ASR}), \mathcal{M}\_b(\text{Ground})) \end{equation}
\begin{equation} \text{Sim\_s} = \cos(\mathcal{M}\_s(\text{ASR}), \mathcal{M}\_s(\text{Ground})) \end{equation}
where $\mathcal{M}\_$ denotes the LLM's hidden states. We then select the \textbf{lower} similarity as the final $\text{Sim}$ score since we intend to capture the semantic gap between the ASR and the ground truth. This embedding-based approach enables a robust evaluation of semantic preservation in ASR outputs, beyond simple word-matching metrics.

\subsection{Apply: QA Accuracy}
\label{annotation error}
The Apply level evaluates a LLM$_\text{Voice}$’s ability to leverage transcribed information for solving downstream problems, reflecting its real-world utility in task-oriented scenarios. We simulate this utility by constructing multi-choice QA, which is a typical listening test for our human language learners. Specifically, for each speech example, we construct $3$ questions based on the ground-truth transcription along with $5$ choices per question. (including $1$ option as ``None of the above'') 

During QA generation, we leverage GPT-4o and prompt it to focus on either the core concept or information details in the transcription. (Appendix~\ref{generateqa})
To mitigate potential errors in generated QA, we also employ GPT-4o itself and Gemini-1.5-flash to answer these questions using the ground-truth transcription. Questions, where GPT-4o or Gemini-1.5-flash fails to produce correct answers, are discarded and regenerated.
Then during evaluation, cascaded systems answer questions based on ASR transcriptions, while end-to-end systems directly process speech inputs to generate responses without intermediate ASR steps. Each LLM$_\text{Voice}$ answers $5$ times per quesion and use the majority vote as the final answer~\cite{wang2023selfconsistencyimproveschainthought}. QA accuracy is then measured via an exact match and higher accuracy indicates a stronger capability to contextualize and apply spoken content to real-world tasks. 

Moreover, we realize that this QA test can also help recognize potential annotation errors in the benchmarks efficiently, since QA generated by annotation errors will be unanswerable to LLM$_\text{Voice}$, we use this feature to filter out annotation errors in our main experiments and further utilize them to detect hallucinations in Section~\ref{unanswerable set}.

\subsection{Final SIQ Score}

Inspired by standard Raven's progressive matrices~\cite{John2003}, a popular human IQ mechanism, we have three steps to compute SIQ: (1) harder samples have higher scores; (2) global standardization making scores among levels computable; (3) dynamic weight among levels in computing final SIQ.
For each LLM$_\text{Voice}$ model \( j \) and each speech sample \( i \), we first compute raw scores for the three dimensions as $X_{\text{WER}, i, j}^{\text{dim}}$ (taking the negative in subsequent computation), $X_{\text{Sim}, i, j}^{\text{dim}}$, and $X_{\text{Acc}, i, j}^{\text{dim}}$.




\paragraph{Speech Sample Discrimination Weights}
To account for variations in sample difficulty, we introduce ``discrimination weights'' based on inter-model score variance. For each speech sample \( i \), we compute the variance across models:
\begin{equation}
    V_i^{\text{dim}} = \text{Var}([X_{\text{Raw}, i, j}^{\text{dim}} \ \forall j])
\end{equation}
where \( "\text{dim}" \) refers to 3 levels.
Using these variance values, we compute weighted scores:

\begin{equation}
   X_{j}^{\text{dim}} = \frac{\sum_{i=1}^{N} X_{\text{Raw}, i, j}^{\text{dim}} \cdot V_i^{\text{dim}}}{\sum_{i=1}^{N} V_i^{\text{dim}}}
\end{equation}
where $N$ denotes the number of models.
This ensures that speech samples with greater discrimination power have a larger influence.

\paragraph{Global Standardization}
For each dimension, we perform standardization to normalize scores across models:

\begin{equation}
    Z_{j}^{\text{dim}} = \frac{X_{ j}^{\text{dim}} - \mu^{\text{dim}}}{\sigma^{\text{dim}}}
\end{equation}
where \( \mu_{\text{dim}} \) and \( \sigma_{\text{dim}} \) are the mean and standard deviation of all model scores for that dimension.





\paragraph{Dynamic Weight Computation}
To ensure a balanced contribution of each evaluation dimension, we assign dynamic weights based on variance. Since higher variance may indicate greater instability, we use the ``inverse variance'':

\begin{equation}
    w^{\text{dim}} = \frac{1}{\sigma^{\text{dim}}_{\text{raw}} + \epsilon}, w_{f}^{\text{dim}} = \frac{w^{\text{dim}}}{\sum w^{\text{dim}}}
\end{equation}
where \( \sigma^{\text{dim}}_{\text{raw}} \) is the standard deviation computed by all raw scores, and \( \epsilon \) is a small constant to prevent division by zero. The weights are then normalized on the summation of three dimensions.


\paragraph{Final IQ Score Computation}
The final intelligence score for each model \( j \) is computed as a weighted sum of the standardized scores:

\begin{equation}
    Score_j = \sum_{\text{dim}} w_{f}^{\text{dim}} \cdot Z^{\text{dim}}_j
\end{equation}
Finally, the score is converted into an IQ-like scale:

\begin{equation}
    SIQ_j = 100 + 15 \cdot Score_j.
\end{equation}

As one additional study, we also perform normalization based on model scale (\textit{i.e.}, analogous to age factors in human IQ). However, due to the involvement of multiple variables in neural scaling laws~\cite{Kaplan2020ScalingLF}, we report the non-normalized SIQ and provide preference rankings from a human study to validate its correlation.


\section{Experiment Setup}
\subsection{Datasets}
To comprehensively evaluate LLM$_\text{Voice}$ intelligence across varied domains and real-world challenges, we carefully select two datasets earning22~\cite{delrio2022earnings22practicalbenchmarkaccents} and voxpopuli~\cite{wang-etal-2021-voxpopuli} from the popular OpenASR Leaderboard, while an additional dataset Med-ASR-EN\footnote{\url{https://huggingface.co/datasets/jarvisx17/Medical-ASR-EN}} is chosen from the medical domain, featuring diverse accents and environments. This ensures that our evaluation tests LLM$_\text{Voice}$ performance in different real-world scenarios, including domain-specific speech and challenging acoustic conditions.
Considering the high computational cost of leveraging GPT-4o APIs for QA evaluation and running end-to-end LLM$_\text{Voice}$ (e.g., Gemini), we extract a subset constructed by the longest audios from each dataset shown in Table~\ref{data:stat}. 
\begin{table}
    \centering
    \resizebox{1.0\linewidth}{!}{
    \begin{tabular}{lrrr}
    \toprule
        Dataset & \# Subset & Domain & WER  \\
        \hline
        Earning22 & 200 & Financial meetings & 12.05 \\
        Voxpopuli & 200 & European Parliament & 7.48\\
        Medasr & 400  & Hospital Accented Patients & 7.7 \\
        \bottomrule
    \end{tabular}
    }
    \caption{\textbf{Statistics of datasets}. WER results are on whole dataset via Whisper-large-v2.}
    \label{data:stat}
\end{table}

\subsection{Models}
We conduct experiments across three major LLM$_\text{Voice}$ architectures as follows:
\paragraph{ASR+LLM} We compare the following widely used ASR models: Whisper-Large-v2, Whisper-Large-v3 \citep{radford2022robustspeechrecognitionlargescale}, Canary \citep{puvvada2024moreaccuratespeechrecognition}, and ESPnet (owsm\_ctc\_v3.1\_1B) \citep{watanabe2018espnetendtoendspeechprocessing}
\paragraph{ASR+GER+LLM} We select GPT-4o \citep{openai2024gpt4ocard} as the GER module and use Whisper-large-v2 to generate 5 hypotheses for each audio based on beam search.
\paragraph{End-to-end Multimodal Models} For speech-to-text multimodal models, we compare Salmonn \citep{tang2024salmonngenerichearingabilities}, Qwen2-audio \citep{chu2024qwen2audiotechnicalreport}, desta2 \citep{lu2025desta2developinginstructionfollowingspeech} and the latest Qwen-2.5-Omni\citep{xu2025qwen25omnitechnicalreport}, we also involve any-to-any multimodel models with speech-to-text capabilities: AnyGPT \citep{zhan2024anygptunifiedmultimodalllm}, Baichuan-omni-1.5 \citep{li2025baichuanomni15technicalreport}, and Gemini-1.5 \citep{geminiteam2024gemini15unlockingmultimodal}. We specifically denote Baichuan-omni-1.5 and Gemini-1.5 as ``end-to-end-large'' since they are either scaled up in training data or model size.

In the understanding level test, we select LLaMA-3.1-8B-Instruct to generate hidden states as responses since related work \citep{jiang-etal-2024-scaling,inproceedings} has proved the strong embedding capability of LLaMA-based models. 

Meanwhile, for the LLM in both ASR + LLM and ASR + GER + LLM to answer QA, we use Qwen2-7b \citep{yang2024qwen2technicalreport} for fair comparisons in size. Preliminary experiments to validate Qwen2-7b and other experiment details are in Appendix~\ref{appendix: experiment}.
\begin{figure}[ht!]
  \includegraphics[width=\linewidth]{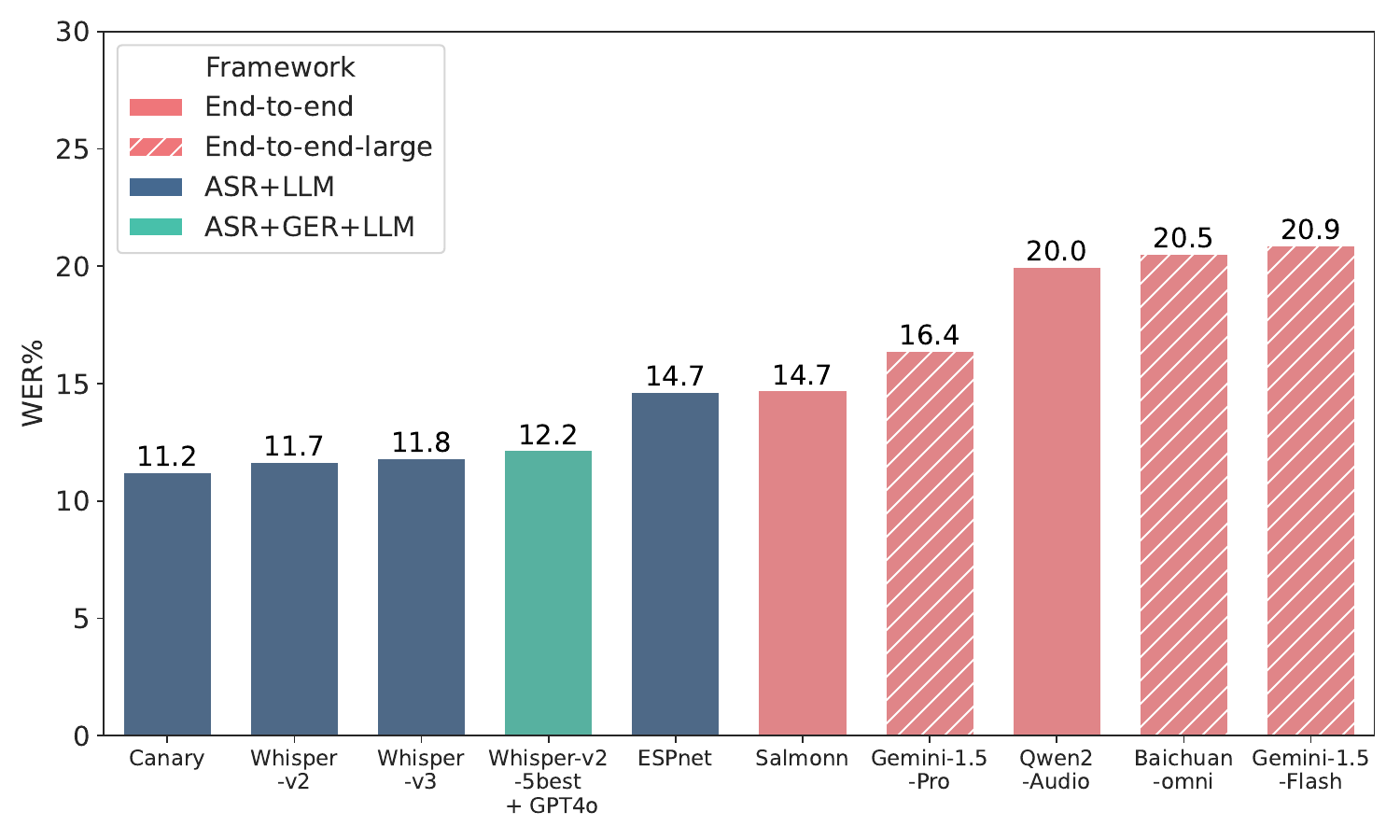} \hfill
  \caption {\textbf{Raw remember scores}. All cascaded models outperform end-to-end models including Gemini-1.5.}
  \label{result: WER}
\end{figure}

\section{Experimental Results}
In this section, we will first show the results of each level, and then the final SIQ scores. Considering the limited space, for each level we show the average score on 3 datasets without Desta2 and AnyGPT since they show a relatively low performance, the complete results are in Appendix.

\begin{figure}[ht!]
  \includegraphics[width=\linewidth]{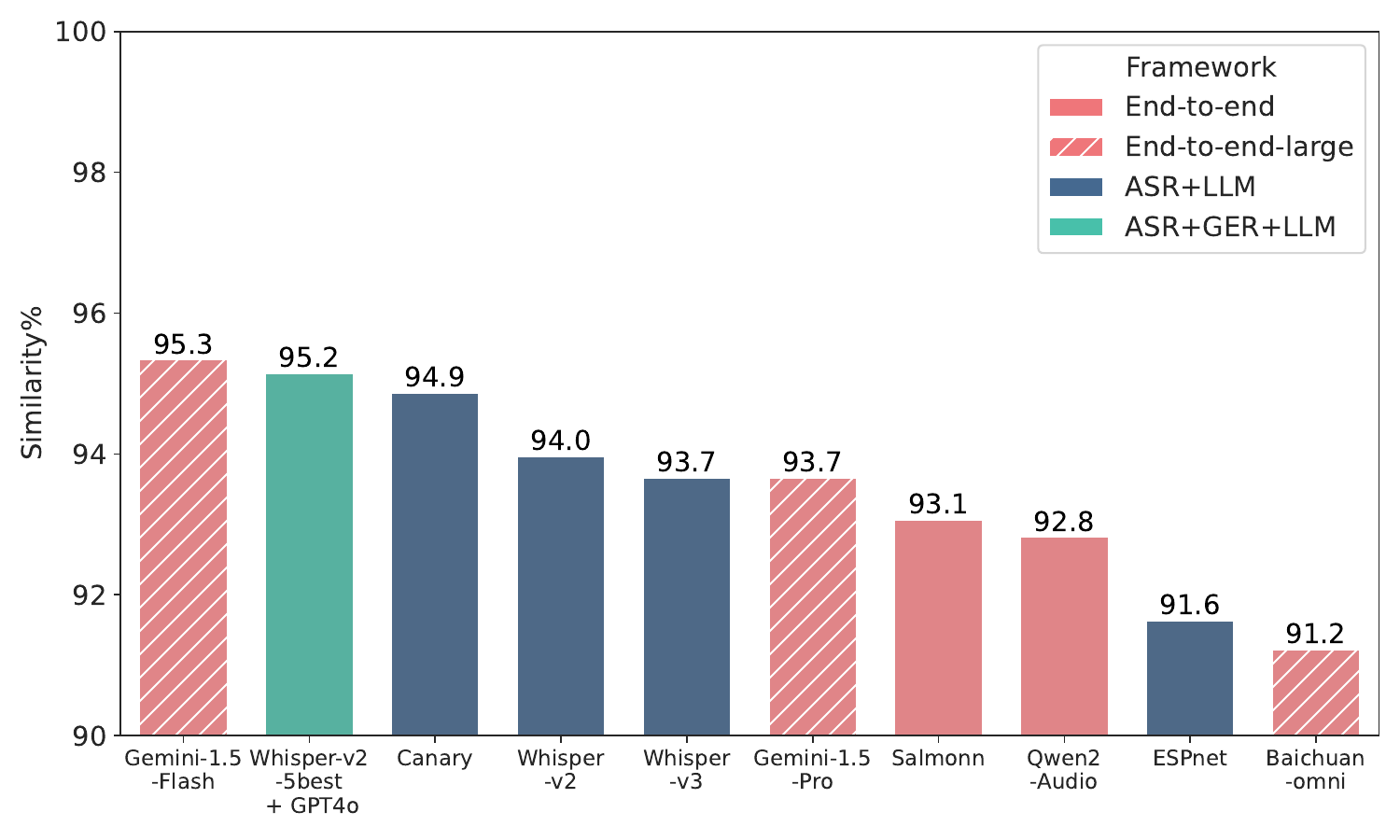}
  \caption {\textbf{Raw understand scores}. Gemini-1.5-flash shows the best semantic capturing.}
  \label{result: understand}
\end{figure}

\subsection{Remember-Level Results}
Figure~\ref{result: WER} shows the raw WER scores of each model. Among all tested models: (1) Canary achieves the lowest WER, demonstrating the best remembering capability; 
(2) GER even negatively affects ASR if we focus on the increasing WER; 
(3) More broadly, all ASR models significantly outperform end-to-end LLM$_\text{Voice}$ in terms of WER, and even large-scale end-to-end models, such as Gemini-1.5, perform worse than ASR models, reinforcing the robustness of traditional dedicated ASR models.

\begin{figure}[ht!]
  \includegraphics[width=\linewidth]{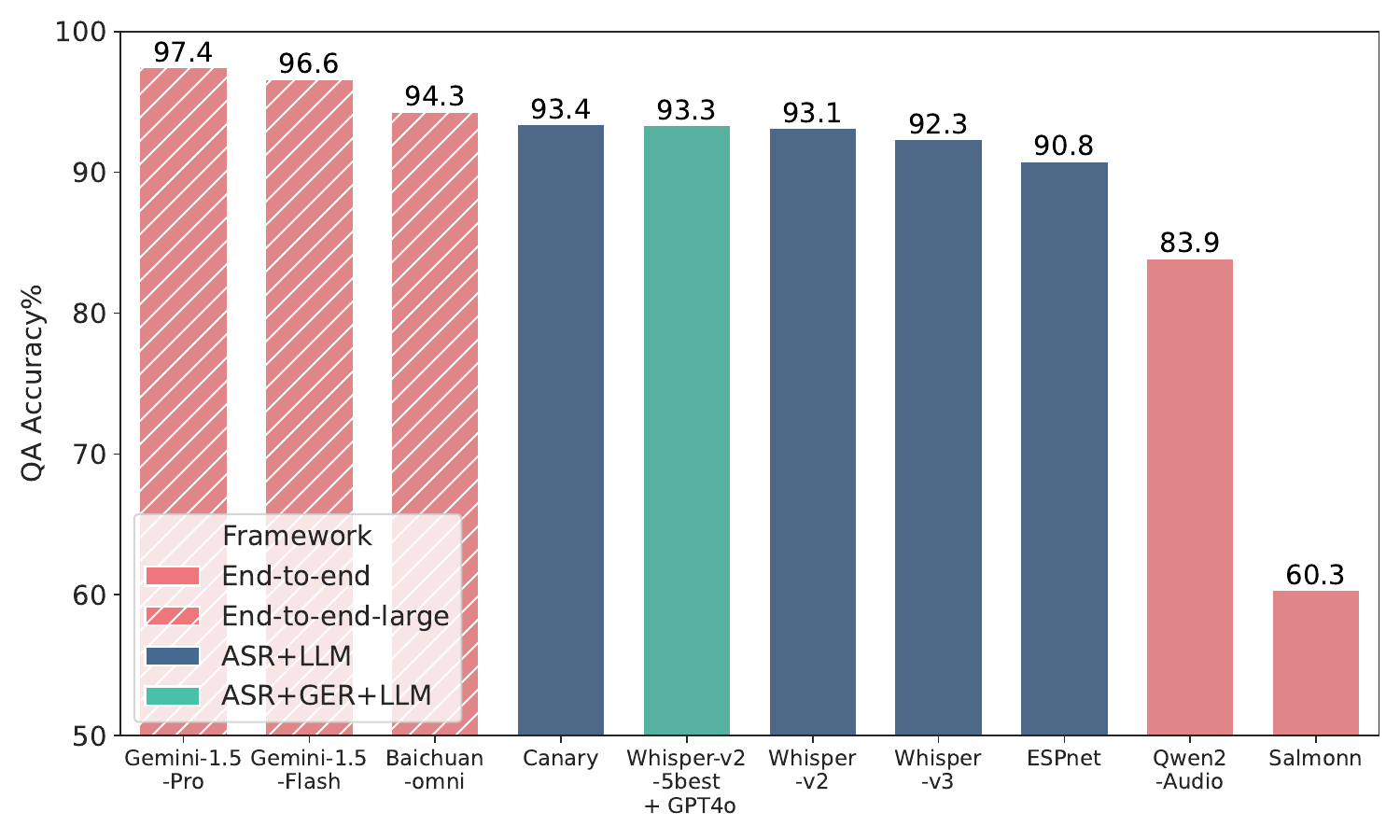}
  \caption {\textbf{Raw apply scores}. End-to-end-large models perform best while smaller ones perform worst.}
  \label{result: apply}
\end{figure}
\begin{table*}[t]
    \centering
   \resizebox{1.0\linewidth}{!}{
    \begin{tabular}{lrrrr}
    \toprule
        Relative Bloom’s Taxonomy Levels & \;\;\;\;\cellcolor[HTML]{FFFC9E}Remember $\uparrow$ \;\;\;\, &  \;\;\;\;\cellcolor[HTML]{96FFFB}Understand $\uparrow$ \;\;\;\, & \;\;\; \cellcolor[HTML]{9AFF99}Apply $\uparrow$ \;\;\;\, & \;\;\;\cellcolor[HTML]{EFEFEF}SIQ $\uparrow$ \;\;\;\,\\

         \toprule
         \multicolumn{5}{c}{\textit{ASR+LLM Approaches}} \\
         \hline
         Whisper$_{\text{v2-1.5B}}$ + Qwen2$_{\text{7B}}$& 0.554 \;\;\;\;\;\,   & 0.499 \;\;\;\;\;\, & 0.481\;\;\;\;\;\, & 107.43\;\;\;\;\;\, \\
         Whisper$_{\text{v3-1.5B}}$ + Qwen2$_{\text{7B}}$& 0.553 \;\;\;\;\;\,  & 0.433 \;\;\;\;\;\, & 0.432\;\;\;\;\;\, & 106.49\;\;\;\;\;\, \\
         Canary$_{\text{1B}}$ + Qwen2$_{\text{7B}}$& \textbf{0.559} \;\;\;\;\;\,  & 0.566 \;\;\;\;\;\,  & 0.504\;\;\;\;\;\,   & 107.78\;\;\;\;\;\, \\
         OWSM-CTC$_{\text{v3.1-1B}}$ + Qwen2$_{\text{7B}}$& 0.534 \;\;\;\;\;\,   & 0.151 \;\;\;\;\;\, &0.353\;\;\;\;\;\, &103.05\;\;\;\;\;\, \\
         \bottomrule
         \multicolumn{5}{c}{\textit{ASR+GER+LLM Approach}} \\
         \hline
        
         Whisper$_{\text{v2-1.5B}}$ + GPT-4o + Qwen2$_{\text{7B}}$ & 0.543 \;\;\;\;\;\,  & 0.632\;\;\;\;\;\,  & 0.487\;\;\;\;\;\, & \textbf{108.64} \;\;\;\;\;\, \\
         \hline
         \toprule
         \multicolumn{5}{c}{\textit{Multi-Modal End-to-End Approaches}} \\
         \hline
         Qwen2-Audio$_{\text{7B w/ 1.5B Whisper}}$ &-0.187\;\;\;\;\;\, & 0.366\;\;\;\;\;\, & 0.011\;\;\;\;\;\, &  103.88\;\;\;\;\;\, \\
         Qwen2.5-Omni$_{\text{7B w/ 1.5B Whisper}}$ &0.472\;\;\;\;\;\, & 0.410\;\;\;\;\;\, & 0.509\;\;\;\;\;\, &  105.74\;\;\;\;\;\, \\
         Salmonn$_{\text{13B w/ 1.5B Whisper}}$&0.508\;\;\;\;\;\, & 0.381\;\;\;\;\;\, & -1.146\;\;\;\;\;\, &  101.03\;\;\;\;\;\,\\
         Desta2$_{\text{8B w/ 1.5B Whisper}}$&-2.575\;\;\;\;\;\,  & -1.604\;\;\;\;\;\, &  -0.233\;\;\;\;\;\,&79.69\;\;\;\;\;\, \\
         AnyGPT$_{\text{7B}}$&0.314\;\;\;\;\;\,  & -2.718\;\;\;\;\;\, &  -2.893\;\;\;\;\;\,&60.02\;\;\;\;\;\, \\
         \hline
         Baichuan-omni-1.5$_{\text{7B}}$&0.448\;\;\;\;\;\,  & 0.184\;\;\;\;\;\, & 0.546\;\;\;\;\;\,&104.02\;\;\;\;\;\, \\  
         Gemini-1.5-flash&-1.885\;\;\;\;\;\,& \textbf{0.641}\;\;\;\;\;\, & 0.673\;\;\;\;\;\, & 107.85\;\;\;\;\;\, \\
         Gemini-1.5-pro&0.492\;\;\;\;\;\,& 0.409\;\;\;\;\;\, & \textbf{0.710}\;\;\;\;\;\, & 107.08\;\;\;\;\;\, \\
         \bottomrule
    \end{tabular}
   }
    \caption{\textbf{Main Results}. The results of remember and understand levels are from single experiment, while apply level include the majority results of 5-time generations.} 
    \label{table: main results}
\end{table*}

\subsection{Apply-Level Results}

As in Figure~\ref{result: apply}: (1) While end-to-end models with similar sizes as cascaded models make much poorer accuracies, all scaling-up end-to-end models achieve superior performance in capturing apply level knowledge, which is consistent with the scaling law; (2) GER not only improves semantic retention but also enhances downstream reasoning capabilities in the apply level. 


\subsection{SIQ Scores}
Table~\ref{table: main results} shows the final SIQ scores and we can observe: 
(1) A key observation is that the rankings based on WER do not hold for overall LLM$_\text{Voice}$' intelligence: while single ASR models outperform ASR+GER and end-to-end approaches on WER, they fail to maintain their lead in higher-level intelligence evaluations;
(2) End-to-end LLM$_\text{Voice}$ systems underperform cascaded models of the same scale, but when model size increases as Gemini-1.5, they achieve competitive SIQ with cascaded approaches;
(3) GER provides stable SIQ score improvements for ASR+GER+LLM models and achieve the best SIQ.
\begin{table}[t]
    \centering
    \resizebox{1.0\linewidth}{!}{
    \begin{tabular}{lrrrrrrr}
    \toprule
        LLM$_{\text{Voice}}$  &\cellcolor[HTML]{EFEFEF}Human&WER&SemDist & LLM-S & BLEU &\cellcolor[HTML]{EFEFEF} SIQ$_\text{rm}$ & \cellcolor[HTML]{EFEFEF} SIQ$_\text{all}$ \\
        \hline
        Model$_{A}$ &3&2&4 & 4 & 3 &4 & 2\\
        Model$_{B}$ &8&10&8 & 8 & 6 &8 &10\\
        Model$_{C}$ &7&7&7 & 6 & 8 & 7&7\\
        Model$_{D}$ &1&4&2 & 5 &4 & 3&1\\
        Model$_{E}$ &4&6&5 & 7 & 5 &6 &4\\
        Model$_{F}$ &6&5&6 & 2 & 7 & 5&6\\
        Model$_{G}$ &2&3&1 & 3 & 2 & 2&3\\
        Model$_{H}$ &5&1&3 & 1 & 1 & 1&5\\
        Model$_{I}$ &9&8&10 & 10 & 9 & 10&8\\
        Model$_{J}$ &10&9&9 & 9 & 10 &9 &9\\
        \hline
        correlation $\rho$ ($\uparrow$) &-&0.770&0.721 & 0.624 & 0.806 & 0.830&\textbf{0.952}\\
        $p$-value ($\downarrow$) &-&0.009&0.005 & 0.019 & 0.054 & 0.003&\small{\textbf{0.00002}}\\
        \bottomrule
    \end{tabular}
    }
    \caption{Human Voted Rankings \& Existing Metrics.}
    \label{result: human rank}
\end{table}


\begin{table}[t]
    \centering
    \resizebox{1.0\linewidth}{!}{
    \begin{tabular}{lcccr}
    \toprule
        Model  &WER $\downarrow$ &Similarity $\uparrow$ &Accuracy $\uparrow$& Hallucination $\downarrow$\\
        \hline
        Whisper$_{\text{v2-1.5B}}$ + Qwen$_{\text{7B}}$ & 11.8&93.7& 91.7 & 2 (\small{12.0\%})\\
        Qwen2-Audio$_{\text{7B}}$ &20.0&92.8& 83.9 & 4 (\small{24.0\%})\\
        \hline
        Whisper$_{\text{v2-1.5B}}$ + Llama3$_{\text{8B-ins}}$ &11.7&94.0& 95.7 & 7 (\small{41.2\%})\\
        Desta2$_{\text{8B w/ 1.5B Whisper}}$ &76.55&81.6& 83.9 &12 (\small{70.6\%}) \\
        \hline
        Whisper$_{\text{v2-1.5B}}$ + Vicuna$_{\text{13B-1.1}}$ &11.7&94.0& 87.3  & 1 (\small{6.8\%})\\
        Salmonn$_{\text{13B w/ 1.5B Whisper}}$ & 14.70&93.1&60.3& 4 (\small{24.0\%}) \\
        \bottomrule
    \end{tabular}
    }
    \caption{\textbf{Cascaded v.s. End-to-end including hallucination detection}. Cascaded models perform better correspondingly. Desta2 shows unexpectedly high QA accuracy and halluciantions along with its foundation Llama3$_{\text{8B-ins}}$.}
    \label{result: ablation}
\end{table}

\paragraph{Human Evaluation on Comparing with Existing Metrics}
To validate that SIQ better reflects the voice understanding capabilities of LLM$_\text{Voice}$ models compared to existing metrics, we conduct a human evaluation on nine examples involving ten randomly selected anonymous models to avoid human bias on models. For each example, we invite ten human experts, including both native speakers and non-native speech researchers, to rank the ten transcriptions produced by each model based on the ground-truth transcription. We then aggregate these rankings to derive a final rank for each model and compare the results with the rankings obtained from existing metrics on the same nine examples. For SIQ, we compare two variants: (1) SIQ$_\text{rm}$: compute on $9$ examples but removing weight standardization due to the limited sample size; (2) SIQ$_\text{all}$: the SIQ in Table~\ref{table: main results} . Then we compute Spearman’s Rank Correlation Coefficient $\rho$ with human rankings. In Table~\ref{result: human rank}, SIQ consistently outperforms existing metrics in capturing human preferences, particularly SIQ$_\text{all}$, demonstrating its effectiveness in assessing voice understanding capabilities.

\subsection{Cascaded vs End-to-end}
Our SIQ scores provide strong empirical evidence that cascaded systems achieve significantly higher intelligence scores compared to end-to-end models of similar size.
However, a crucial factor influencing this result is the LLM used in QA. Cascaded systems can typically utilize the latest LLMs (e.g., Qwen2-7b, GPT-4o) for QA-based reasoning, whereas end-to-end models are often constrained by their foundation LLMs that jointly handle speech tokens and language understanding. 
To eliminate this confounding factor, we conducted an ablation experiment comparing cascaded and end-to-end systems built on the same base model. 



As shown in Table~\ref{result: ablation}, the cascaded approach consistently outperforms the end-to-end model in 3 levels. Specifically, Qwen2-audio and Salmonn are largely worse than their cascaded variants at the apply level. We hypothesize that the multi-modal training may sacrifice the original reasoning capability of foundation models if not well designed, which is crucial for understanding the limitations of joint training for multi-modal intelligence.

Additionally, we observe that Desta-2 exhibits unexpectedly high QA accuracy while its performances on the WER and similarity are pretty poor. Interestingly, the base model of Desta-2, LLaMA3-8B-instruct, also achieves a higher-than-expected performance than cascaded experiments. We implemented a closer case study (Appendix~\ref{case study}) which reveals that Desta-2 may stem from LLaMA3-8B-instruct showing a strong tendency to hallucinate: (1) Guess the answer even not been recognized in its ASR results; (2) Change the options in the question. This indicates that SIQ can not be represented by the apply level alone, which may suffer from the hallucination issue in question answering. 

Moreover, manually verifying hallucinated QA evaluation is prohibitively expensive. Inspired by the ``annotation error'' introduced in Section~\ref{annotation error}, in the following section, we introduce an unanswerable set to detect hallucinations in LLM$_\text{Voice}$.

\subsection{Unanswerable Set: Detecting Hallucination in LLM$_\text{Voice}$}\label{unanswerable set}
One major advantage of QA-based evaluation is that it allows us to efficiently identify potential annotation errors. If a given QA pair consistently fails across most LLM$_\text{Voice}$, it is likely that the failure is due to incorrect or ambiguous annotations resulting in unanswerable questions rather than genuine model errors.

To leverage this property, we filter out questions that the majority (a threshold of "half") of LLM$_\text{Voice}$ fail to answer correctly and manually check the corresponding speech and annotations.
This significantly reduces human efforts overhead (only $26$ out of $800$ samples require human review). After verification, we confirm that $17$ questions are truly unanswerable based on the available speech content, which forms our unanswerable Set for detecting hallucinations.


We measure both cascaded and end-to-end LLM$_\text{Voice}$ in Table~\ref{result: ablation}. For each unanswerable question, the hallucination will be counted when the answer to the unanswerable question is not ``(E) None of the above,''.
Our results indicate that LLaMA3-8B-Instruct exhibits a significantly higher hallucination ratio than other LLMs, and this problem gets even worse in its end-to-end variant, which explains the higher-than-expected performance in apply level test.
This inspires us that the hallucination of foundation LLMs may be inherited by its multi-modal variances, emphasizing the importance of foundation model selection and reducing hallucination in multi-modal training.

\subsection{End-to-end Models on the Cascaded Category}
We then checked the performance of end-to-end models (Qwen2-audio and Gemini-1.5-flash) in the cascaded category by replacing the ASR model by the end-to-end model, and then used the LLM in the cascaded category to answer questions, noting that this variant only affects the apply level evaluation. Interestingly, in figure~\ref{result: discuss}we find that for the smaller model Qwen2-audio, cascaded version improves the apply level while for larger model Gemini, end-to-end version wins. This may indicate that higher intelligence may result in better end-to-end apply level performance even if sometimes perfect transcription is not generated. We further extend this assumption by involving a human SIQ test, we manually select 3 hard examples and invite 10 human individuals to report the average scores as in Table~\ref{result: humaniq}. It seems that humans are still good at apply level even with the worst WER performance in transcription, which is consistent with our assumption that higher intelligence may largely benefit higher-order cognitive levels. This may inspire future work on how to improve the intelligence level of LLM$_\text{Voice}$.

\begin{table}[t]
    \centering
    \resizebox{1.0\linewidth}{!}{
    \begin{tabular}{lrrr}
    \toprule
        Model  &Remember $\downarrow$ &Understand 
        $\uparrow$& Apply $\uparrow$\\
        \hline
        Qwen2-Audio$_{\text{7B}}$ (end2end) &20.00&92.8&83.9\\
        Qwen2-Audio$_{\text{7B}}$ (cascaded) &20.00&92.8&84.2\\
        \hline
        Gemini-1.5-flash (end2end) &20.9&95.3&96.6\\
        Gemini-1.5-pro (cascaded) &20.9&95.3&95.1\\
        \bottomrule
    \end{tabular}
    }
    \caption{\textbf{Take ASR from end-to-end models}}
    \label{result: discuss}
\end{table}

\begin{table}[t]
    \centering
    \resizebox{1.0\linewidth}{!}{
    \begin{tabular}{lrr}
    \toprule
        Model  &Remember $\downarrow$ &Apply $\uparrow$\\
        \hline
        Whisper$_{\text{v2-1.5B}}$ + Qwen$_{\text{7B}}$ & 23.71&88.89\\
        Whisper$_{\text{v2-1.5B}}$ + Qwen$_{\text{7B}}$ & 17.26&88.89\\
        Canary$_{\text{1B}}$ + Qwen$_{\text{7B}}$ & 24.80&88.89\\
        \hline
        Whisper$_{\text{v2-1.5B}}$ + GPT-4o + Qwen2$_{\text{7B}}$& 21.13&88.89\\
        \hline
        Salmonn$_{\text{13B w/ 1.5B Whisper}}$ &31.55&44.44\\
        Qwen2-Audio$_{\text{7B}}$ &25.50&88.89\\
        Gemini-1.5-flash &19.05&66.67\\
        Gemini-1.5-pro & 19.05&77.78\\
        \hline
        Human & \textbf{53.67}&\textbf{99.27} \\
        \bottomrule
    \end{tabular}
    }
    \caption{\textbf{Human test on 3 examples.}}
    \label{result: humaniq}
\end{table}

\section{Conclusion}
Existing metrics fail to comprehensively evaluate LLM$_\text{Voice}$’s capability in semantic understanding or task-solving. Therefore, 
we build a three-level SIQ test following Bloom’s taxonomy, each level owns unique features and represents different intelligence levels.
Our study involves various LLM$_\text{Voice}$ frameworks, and the SIQ test serves as a unified metric for benchmarking different frameworks. We then discuss whether current end-to-end LLM$_\text{Voice}$s outperform cascaded methods with the same size, indicating the potential insights in training multi-modal models.
Moreover, benefiting from the QA test, we need much less human effort to filter out annotation errors in existing benchmarks and build an unanswerable set, which can further help evaluate the hallucination of LLM$_\text{Voice}$ which is also crucial for multi-modal training.

\section*{Limitation}
In our attempt to provide an examination of Speech IQ, our study limits that warrant discussion.

\paragraph{Data and Evaluation Limitations}
Although this work represents the first systematic investigation of SIQ assessment with demonstrated effectiveness, our current evaluation is based on moderately sized test sets. In future work, we plan to extend our validation framework across diverse domains and languages, and we will employ rigorous quality assurance techniques to filter annotation errors in widely used benchmarks. Moreover, while using publicly available datasets ensures reproducibility, it also introduces potential risks of data leakage. To mitigate this, we are developing closed datasets with controlled knowledge cutoffs for subsequent research phases.

\paragraph{Scaling and Quantitative Analysis}
Our results indicate that SIQ is sensitive to model scaling effects, including variations in architecture size and training data volume. However, our current analysis does not quantitatively characterize these relationships, such as a limitation reminiscent of the challenges in interpreting human IQ with age-group normalization. As discussed in Appendix~\ref{size normalization}, our next SIQ iteration will incorporate scaling law normalization protocols to decouple intrinsic voice understanding capabilities from artifacts induced by parametric scaling, enabling a more nuanced analysis of model performance.

\paragraph{Ethical and Societal Considerations}
The introduction of SIQ raises important ethical and societal questions. Since IQ classification in humans can lead to social discrimination, we are concerned that analogous issues might arise in AI, potentially resulting in biased treatment of systems with lower SIQ scores. We are committed to addressing these concerns proactively and developing safeguards to prevent such discrimination.

\paragraph{Upper 3 Levels in Bloom Taxonomy}
Our study focuses only on the bottom three layers of the Bloom Taxonomy, leaving the upper three layers unexplored. While this approach provides a solid foundation for understanding the core aspects of the problem, it does not capture the full hierarchical structure. The upper layers, which may involve more advanced levels of \textit{analysis}, \textit{evaluation}, and \textit{creation}, remain an open avenue for our future research incorporating audio generation, physical simulation, and acoustic event reasoning models~\cite{pmlr-v235-kong24a}. Expanding the analysis to all 6 layers could provide a more comprehensive understanding of its broader implications toward a form of ``audio intelligence'' for voice assistants. 

  \section*{Acknowledgements}
This work was supported by Cross‑ministerial
Strategic Innovation Promotion Program
(SIP) on ``Integrated Health Care System''
Grant No. JPJ012425 and JSPS KAKENHI Grant
No. JP24KJ1442 and NVIDIA Taiwan Research and Devolvement Center (TRDC) project. 
  
\bibliography{anthology,custom}
\clearpage
\appendix

\section{QA Generation}\label{generateqa}

\begin{figure}[ht!]
  \includegraphics[width=\linewidth]{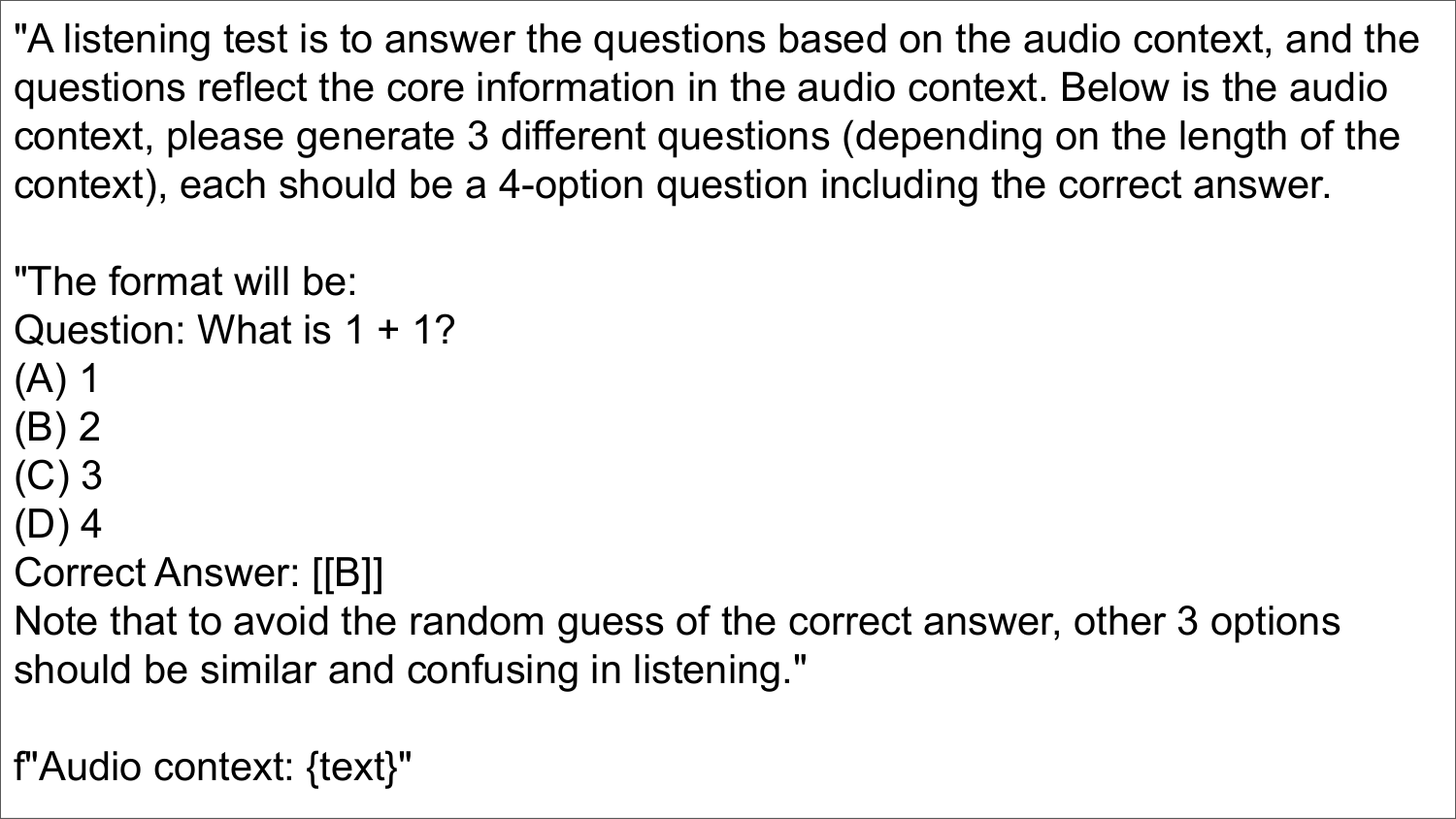}
  \caption{Prompt for QA generation}
  \label{prompt: gpt}
\end{figure}

Since we intend to follow the English test for human learners, our prompt emphasize our intention to instruct GPT-4o is in Figure~\ref{prompt: gpt}. We then add another option ``None of the above'' to each question. Besides, to mitigate potential hallucinations in question generation, we employ a two-step validation process: (1) GPT-4o generates three candidate questions per speech example, and (2) GPT-4o itself and Gemini-1.5-flash attempt to answer these questions using the ground-truth transcription. Questions where GPT-4o or Gemini-1.5-flash fails to produce correct answers are discarded and regenerated, ensuring robust alignment between audio content and QA pairs
\section{Experiment Setup}
\label{appendix: experiment}
We keep the default settings (e.g., temperature) for all models referring their official repositories or huggingface pages.

\begin{table}[ht!]
    \centering
    \resizebox{1.0\linewidth}{!}{
    \begin{tabular}{lrr}
    \toprule
        ASR & Answering LLM & Accuracy  \\
        \hline
        Whisper-large-v2 & GPT-4o & 93.3 \\
        Whisper-large-v2 & Qwen2-7B & 93.1 \\
        Whisper-large-v3 & GPT-4o &  92.3\\
        Whisper-large-v3 & Qwen2-7b &92.3  \\
        Canary & GPT-4o & 93.7 \\
        Canary & Qwen2-7B &  93.4\\
        \bottomrule
    \end{tabular}
    }
    \caption{\textbf{Results of same ASR models with different LLMs in answering QA}.}
    \label{exp:qwen2}
\end{table}
\subsection{Validation of Qwen2-7B}
To ensure the robustness of Qwen2-7B in cascaded approaches, we compared with GPT-4o for several ASR models. As shown in Table~\ref{exp:qwen2}, Qwen2-7B shows a very closer performance with GPT-4o in QA, as well as a similar model size with end-to-end models for fair comparisons.

\subsection{Prompts for end-to-end models}
\begin{figure}
  \includegraphics[width=\linewidth]{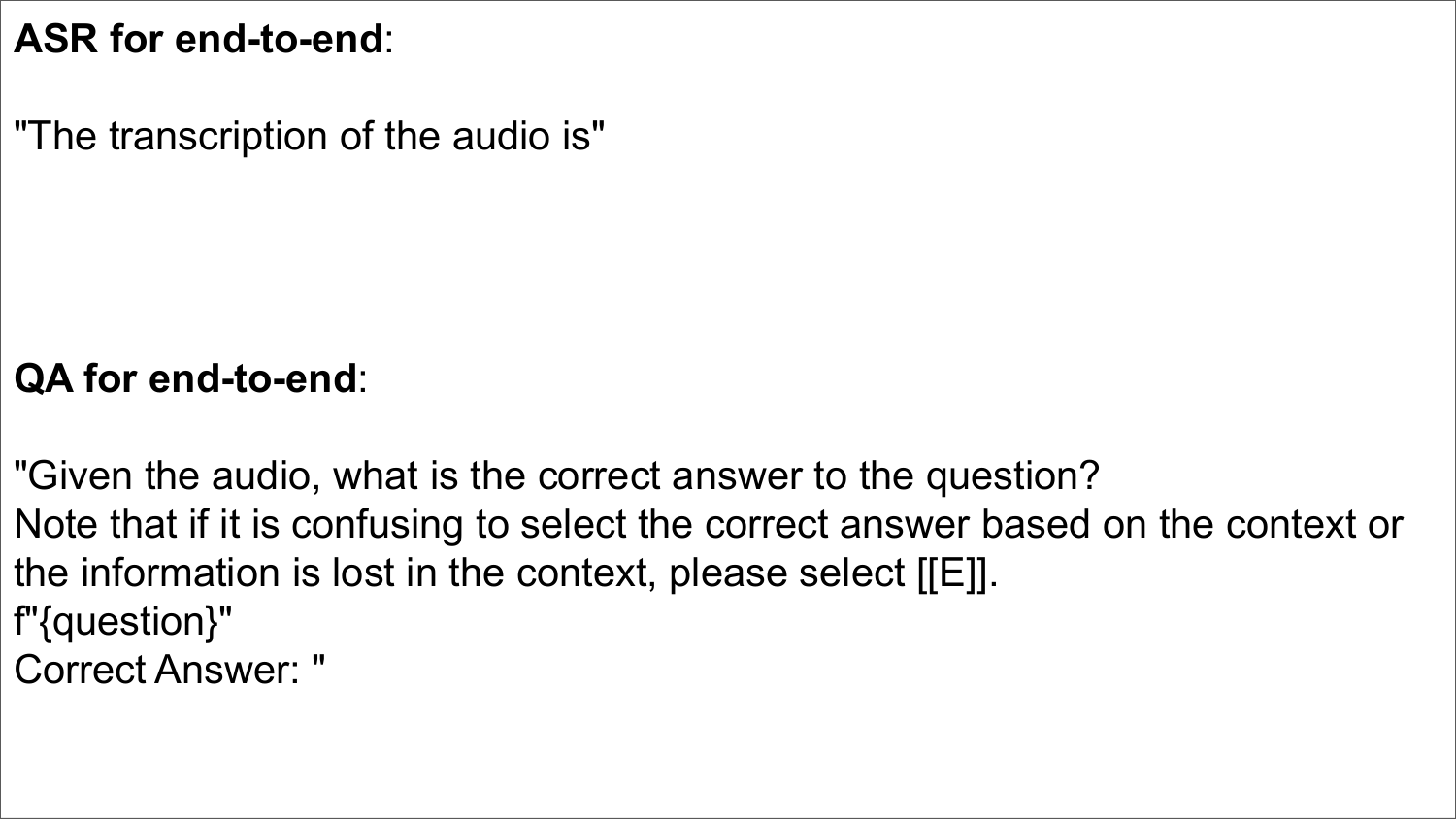}
  \caption{Prompt for end-to-end models}
  \label{prompt: asr}
\end{figure}

Considering that a model-specific prompt is not realistic due to the closed training data for some end-to-end models, we use a unified prompt for instructing ASR transcription as in Figure~\ref{prompt: asr}
\subsection{Prompts for cascaded models in QA}
\begin{figure}
  \includegraphics[width=\linewidth]{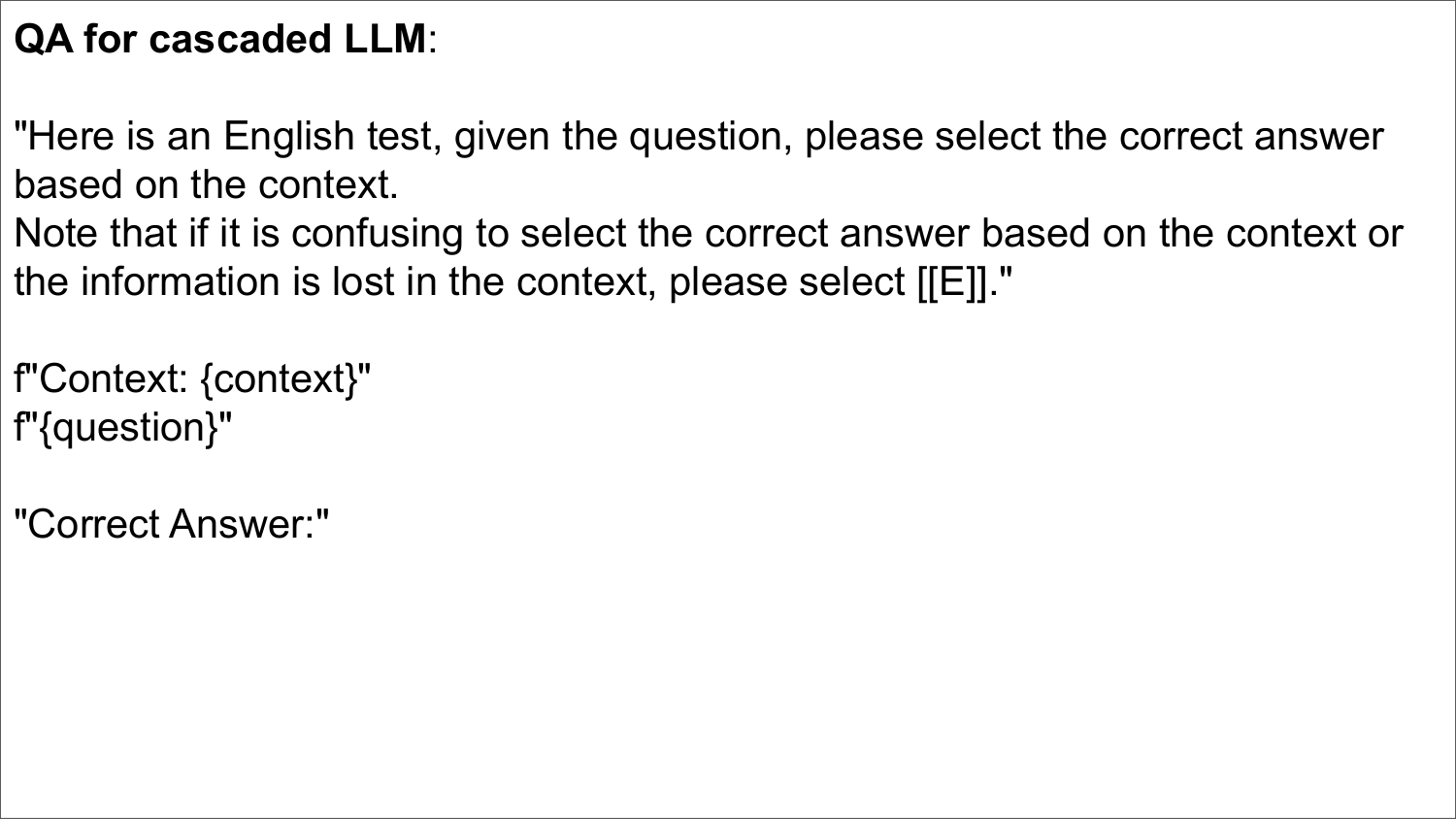}
  \caption{Prompt for QA generation}
  \label{prompt: qa}
\end{figure}

Our prompt for instructing cascaded LLM$_\text{Voice}$s in QA as in Figure~\ref{prompt: qa}.

\section{Case Study}
\label{case study}
\begin{figure*}
  \includegraphics[width=0.48\linewidth]{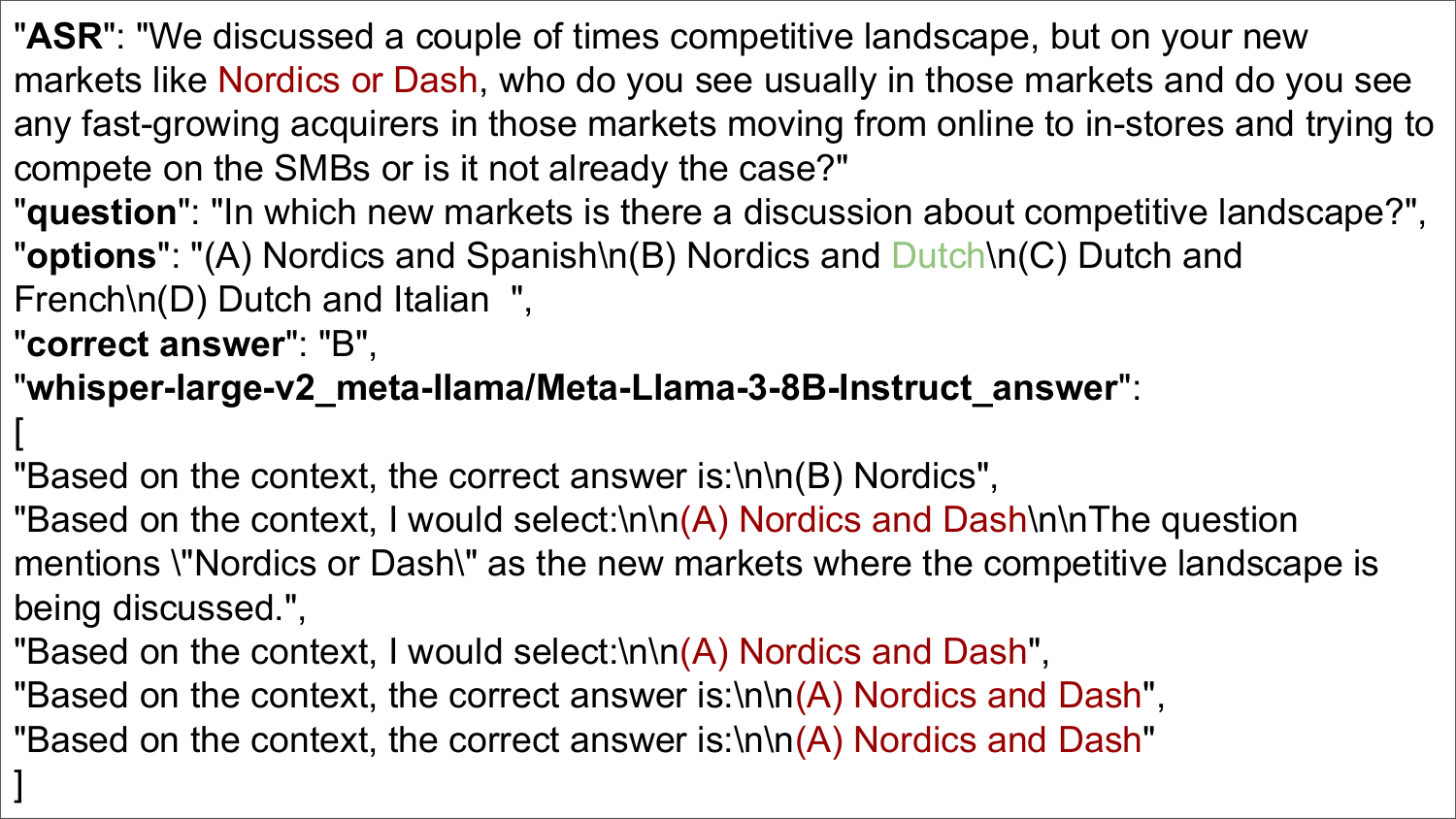} \hfill
  \includegraphics[width=0.48\linewidth]{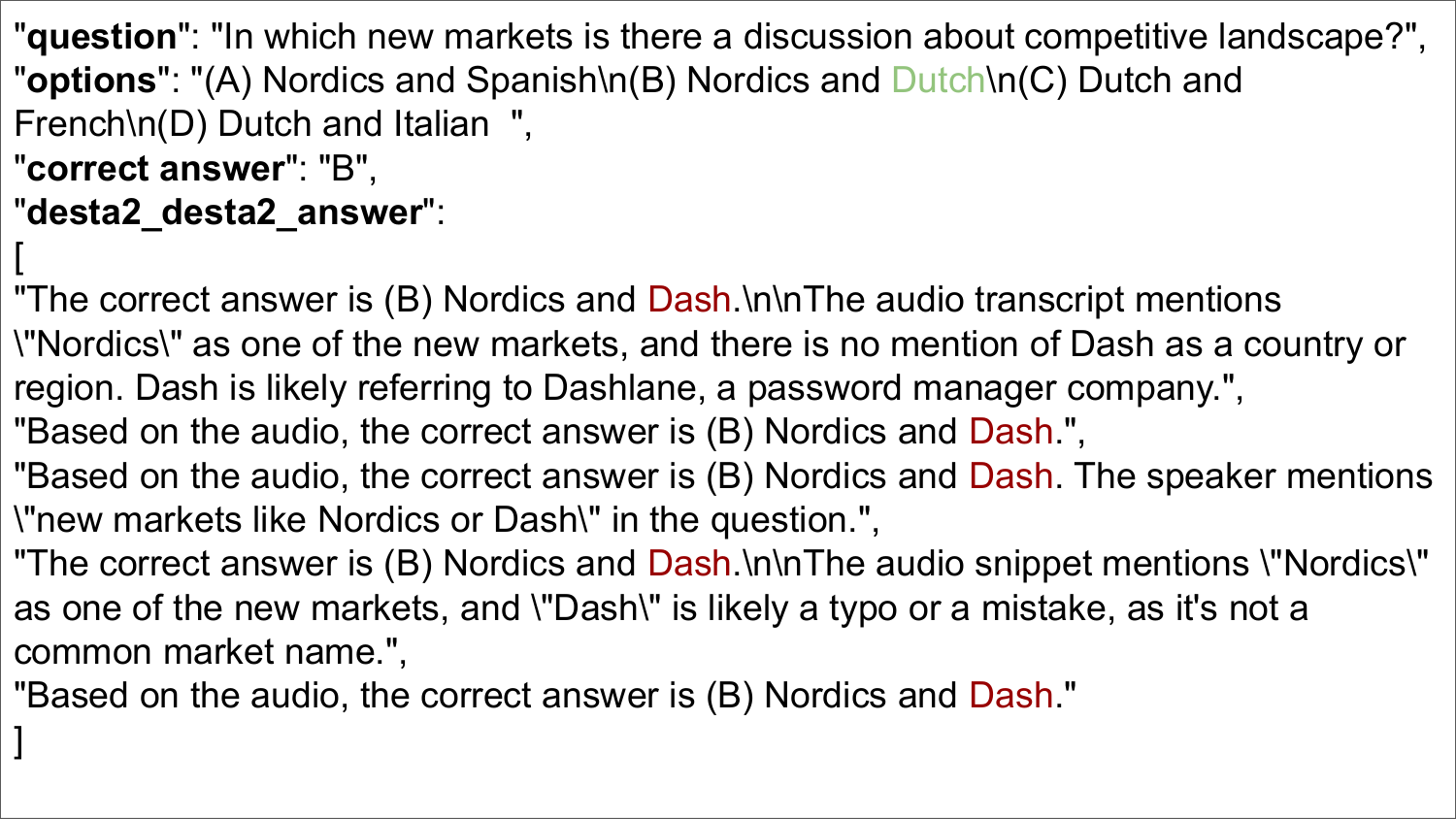}
  \caption {Left: Llama3-8B-Instruct, Right: Desta2. \textcolor{red}{Red} the errors in answering, we find that Llama-8B-Instruct based on the error ASR even hallucinate to change the option in QA, while Desta2 though answer correctly, still inherit this hallucination. }
  \label{case1}
\end{figure*}

\begin{figure*}
  \includegraphics[width=0.48\linewidth]{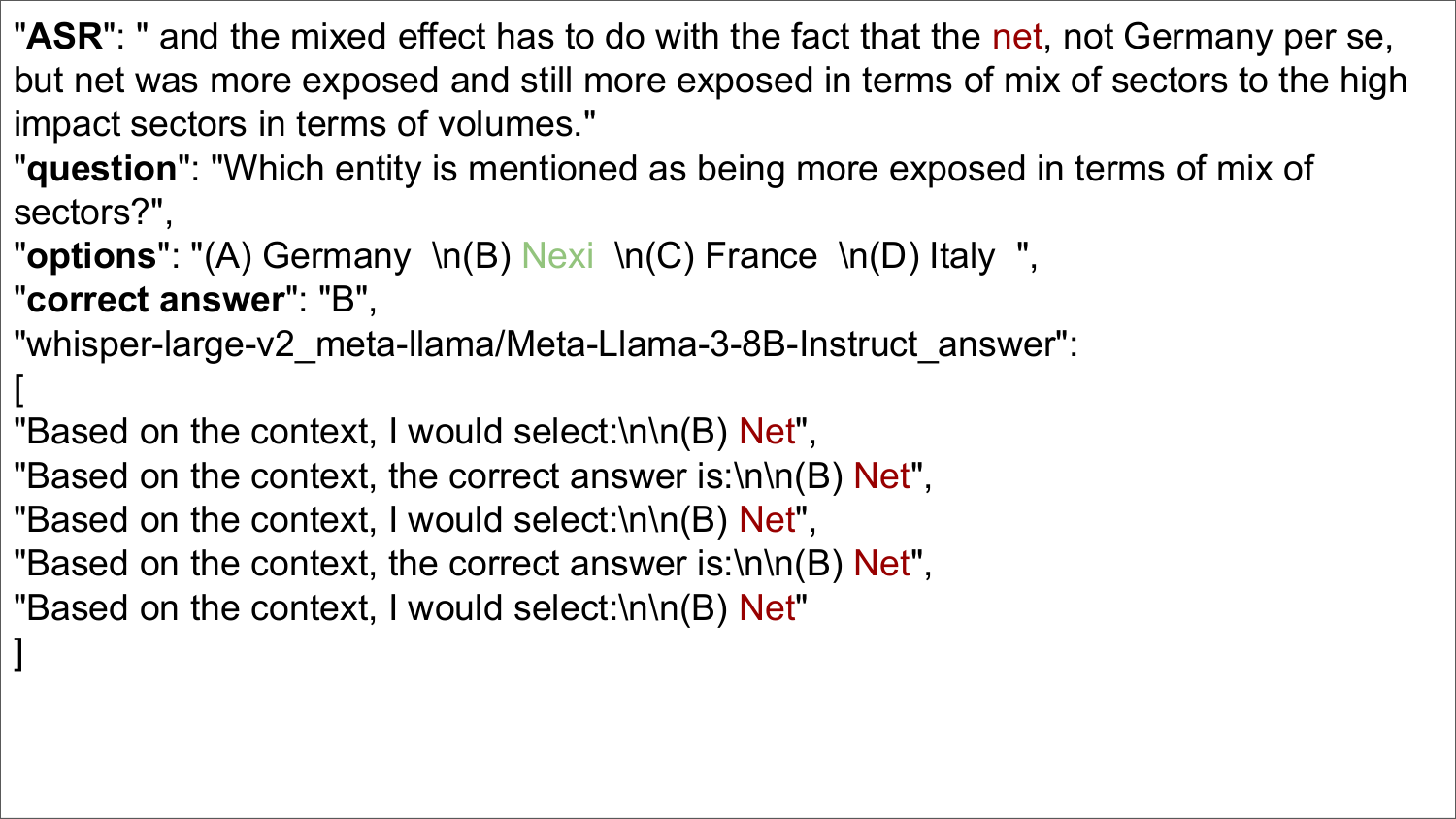} \hfill
  \includegraphics[width=0.48\linewidth]{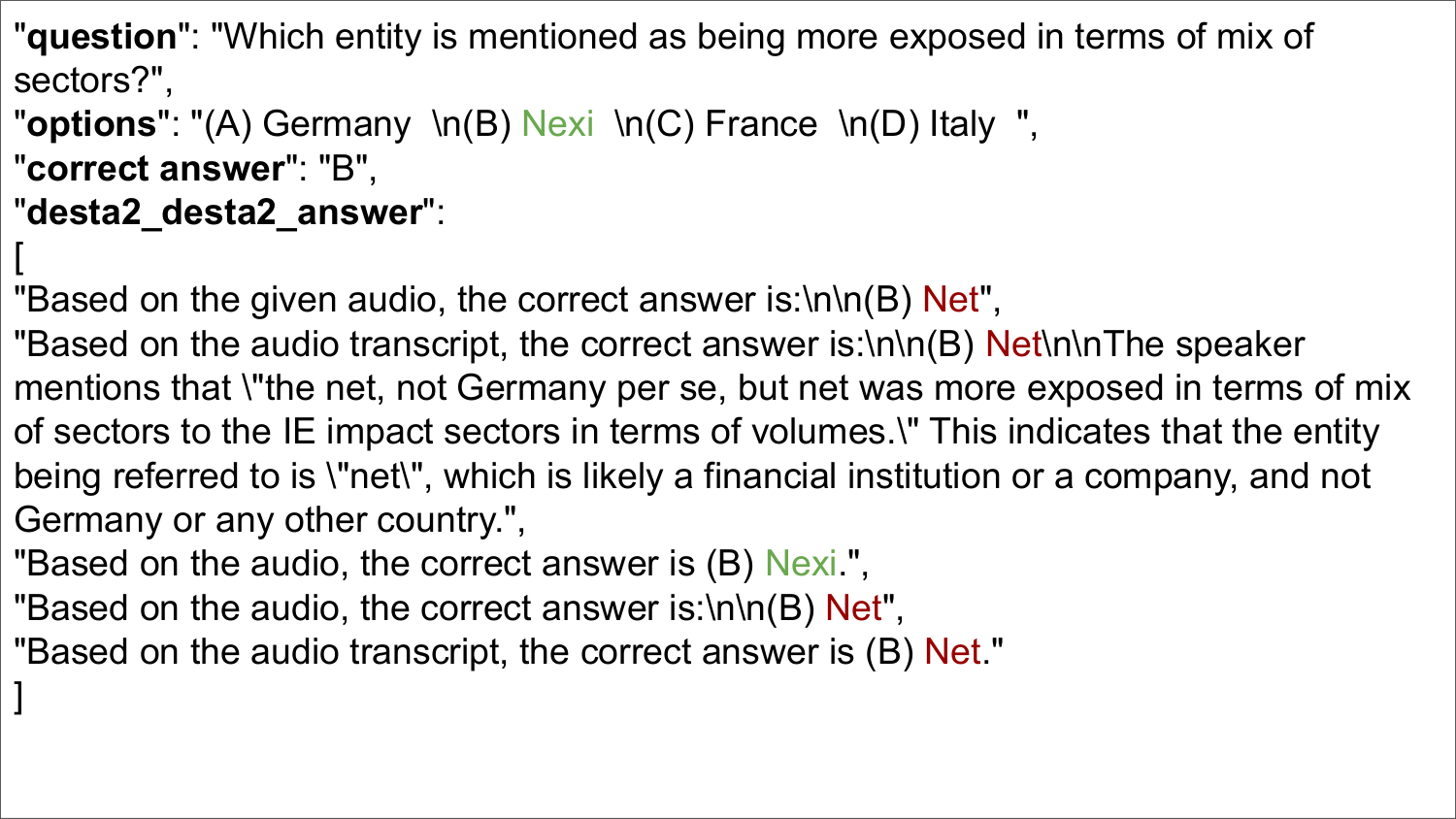}
  \caption {Left: Llama3-8B-Instruct, Right: Desta2. \textcolor{red}{Red} the errors in answering, we find that Llama-8B-Instruct gives the correct answer but based on the error ASR and also hallucinate to change the option in QA, while Desta2 keeps the same hallucination. }
    \label{case2}
\end{figure*}

We manually check the outputs of Desta2 and LLama3-8B-Instruct to investigate the reason behind unexpectedly high QA accuracy. As shown in Figure~\ref{case1}~\ref{case2}, in 5-time generations, the hallucination of Llama3-8B-Instruct lies in either giving a correct answer based on the error ASR or directly changing the options in QA, and interestingly, this halluciantion is kept by its end-to-end variant Desta2. 

\section{Complete Results}
\subsection{WER Results}
\begin{figure}
  \includegraphics[width=\linewidth]{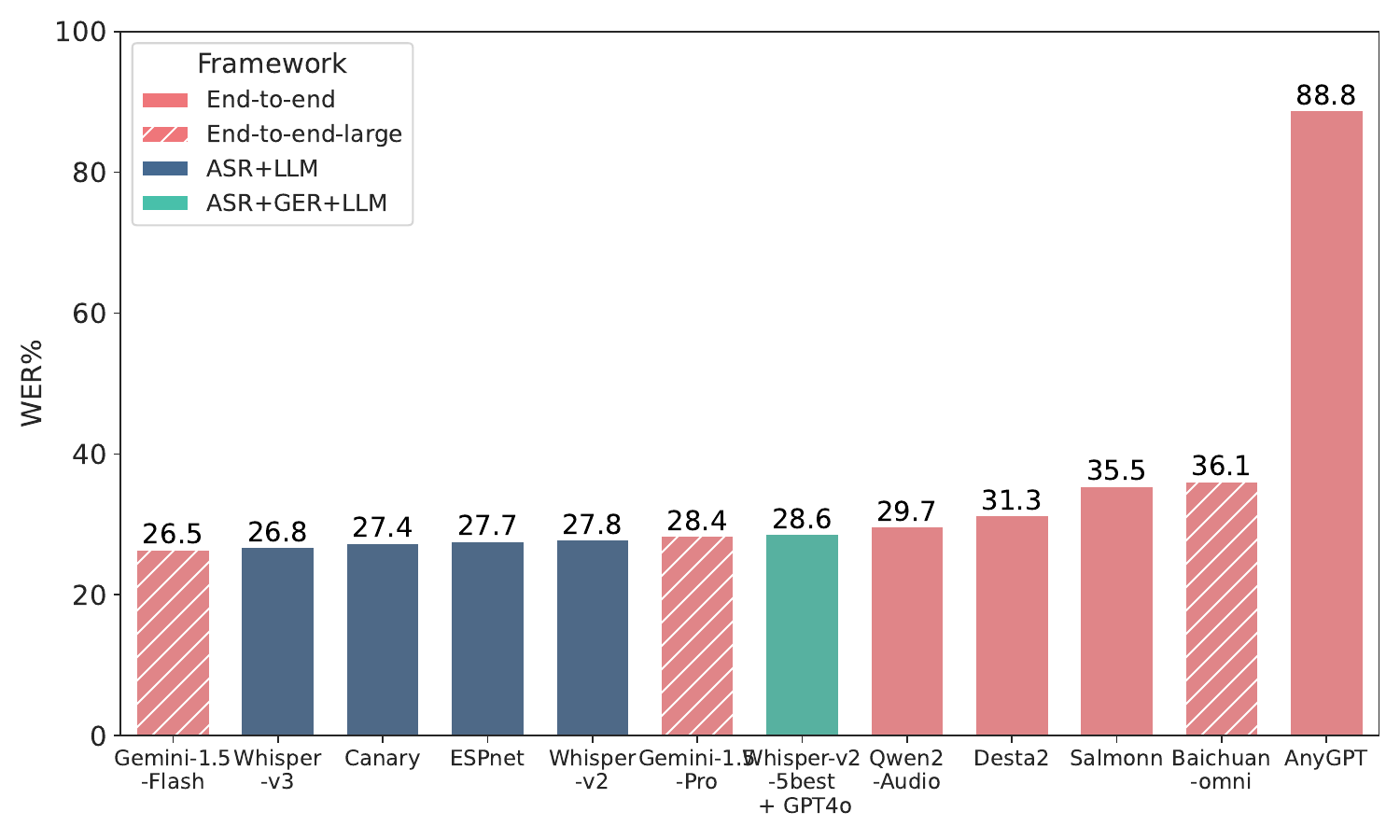}
  \caption{WER on earning22}
  \label{wer_earning22}
\end{figure}

\begin{figure}
  \includegraphics[width=\linewidth]{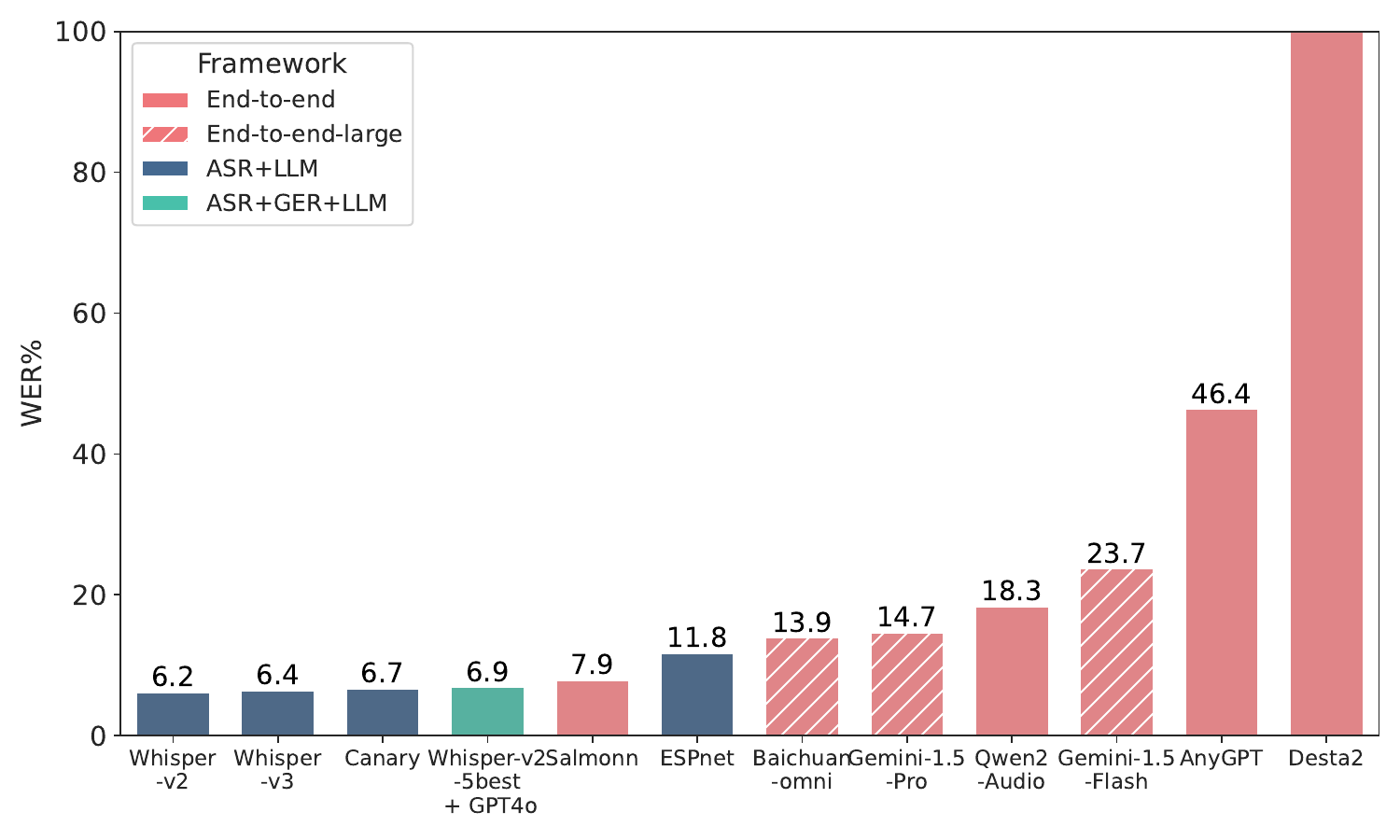}
  \caption{WER on medasr}
  \label{wer_medasr}
\end{figure}

\begin{figure}
  \includegraphics[width=\linewidth]{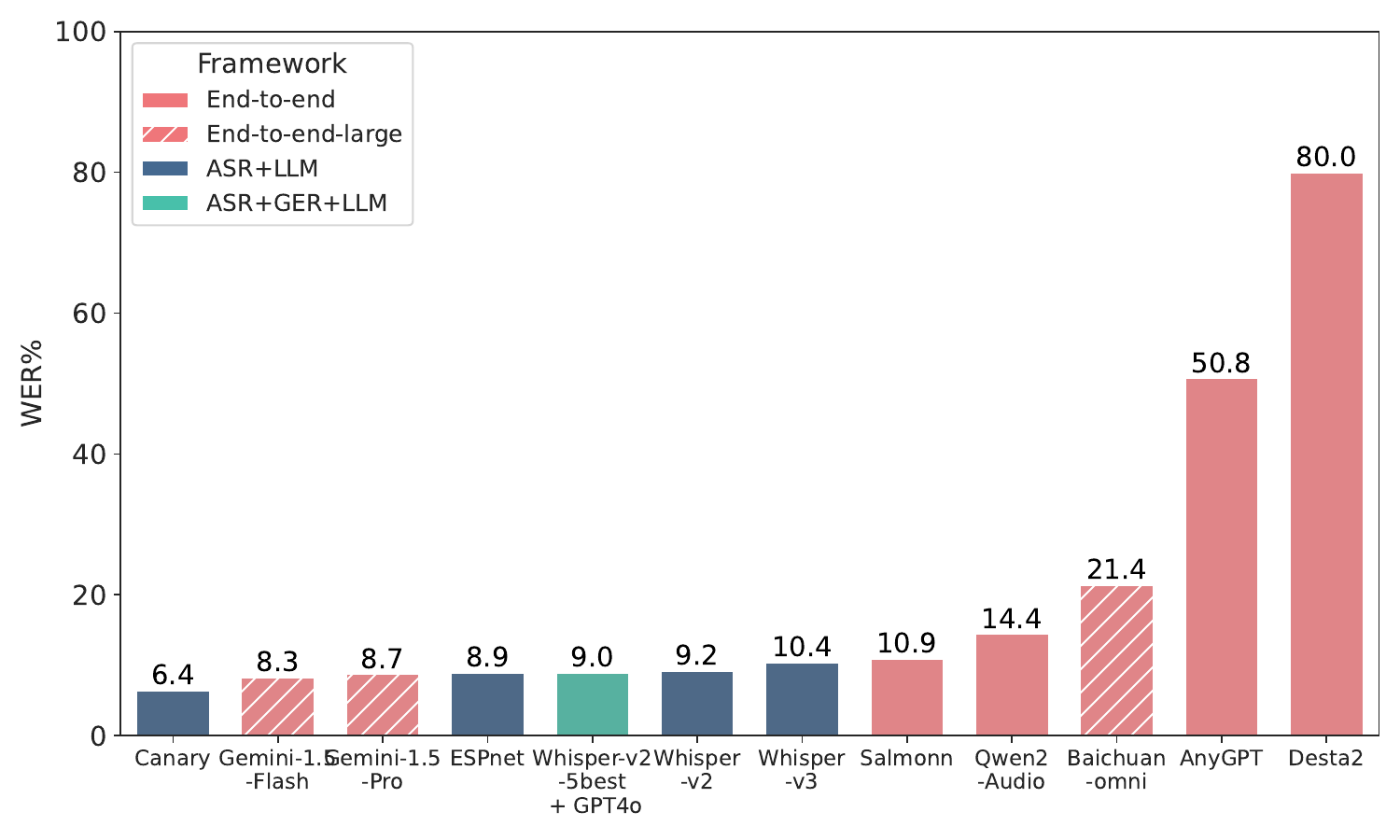}
  \caption{WER on voxpopuli}
  \label{wer_voxpopuli}
\end{figure}

As in Figure~\ref{wer_earning22}~\ref{wer_medasr}~\ref{wer_voxpopuli}, the WER performance on each dataset show a consistent tendency as the average score that cascaded LLM$_\text{Voice}$ overally show better performance than end-to-end LLM$_\text{Voice}$s, and GER seems to negatively influence the ASR. We also notice that AnyGPT and Desta2 show much lower performance than other models, with closer check into their outputs, we find that sometimes they can not recognize the voice and provide meanlingless responses, while their foundation ASR models can succeed in recognizing. This may indicate the potential problem in their modality alignment training.

\subsection{Similarity Results}
\begin{figure}
  \includegraphics[width=\linewidth]{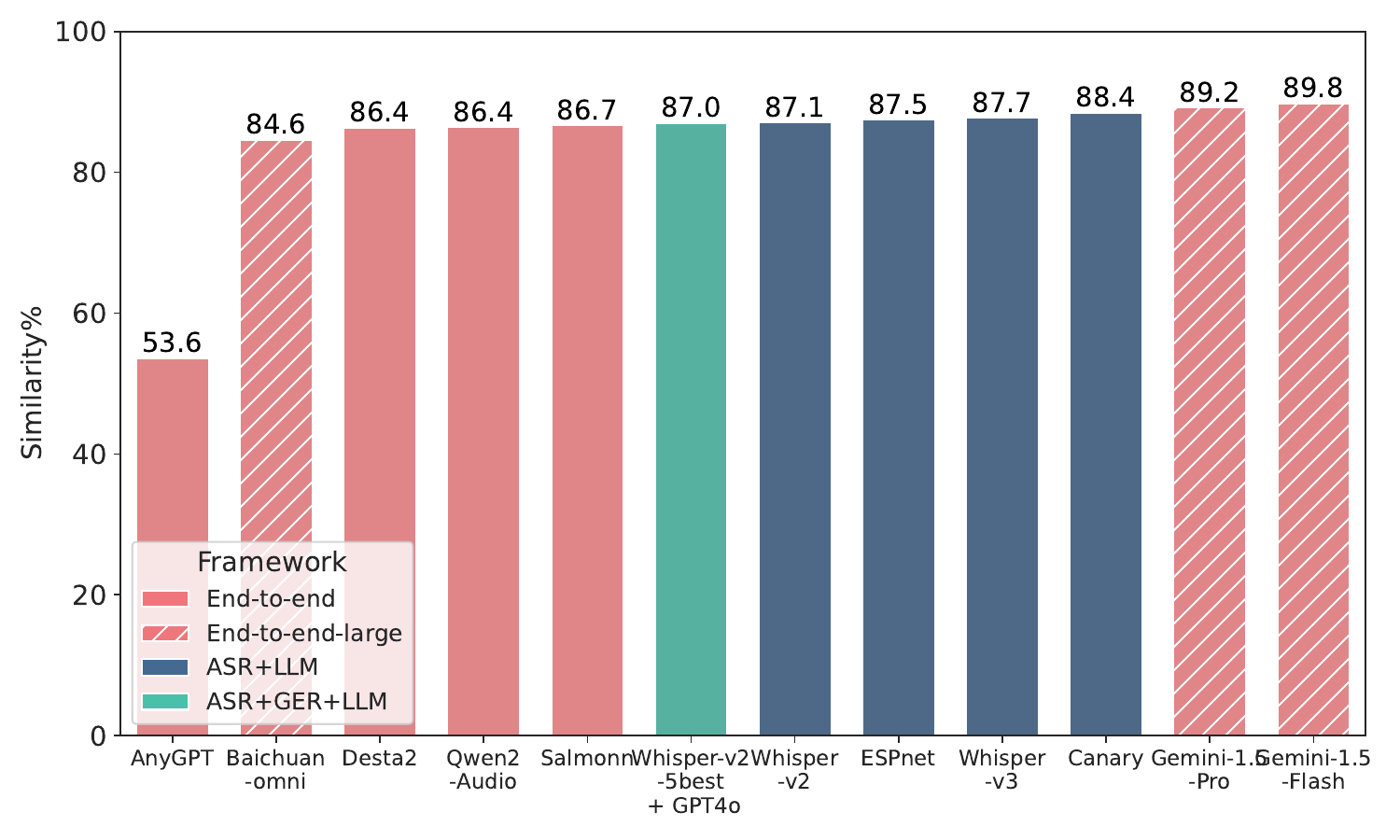}
  \caption{Similarity on earning22}
  \label{sim_earning22}
\end{figure}

\begin{figure}
  \includegraphics[width=\linewidth]{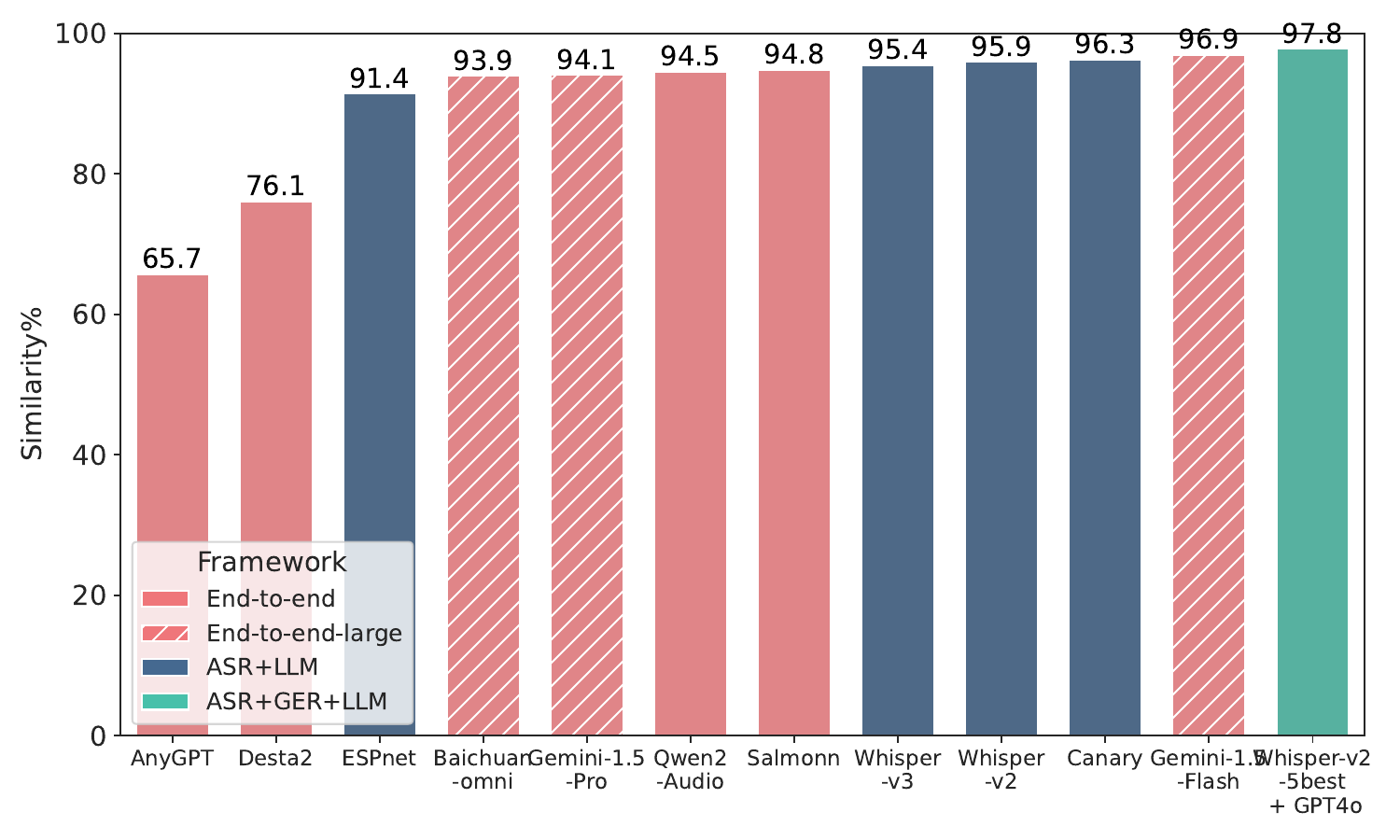}
  \caption{Similarity on medasr}
  \label{sim_medasr}
\end{figure}

\begin{figure}
  \includegraphics[width=\linewidth]{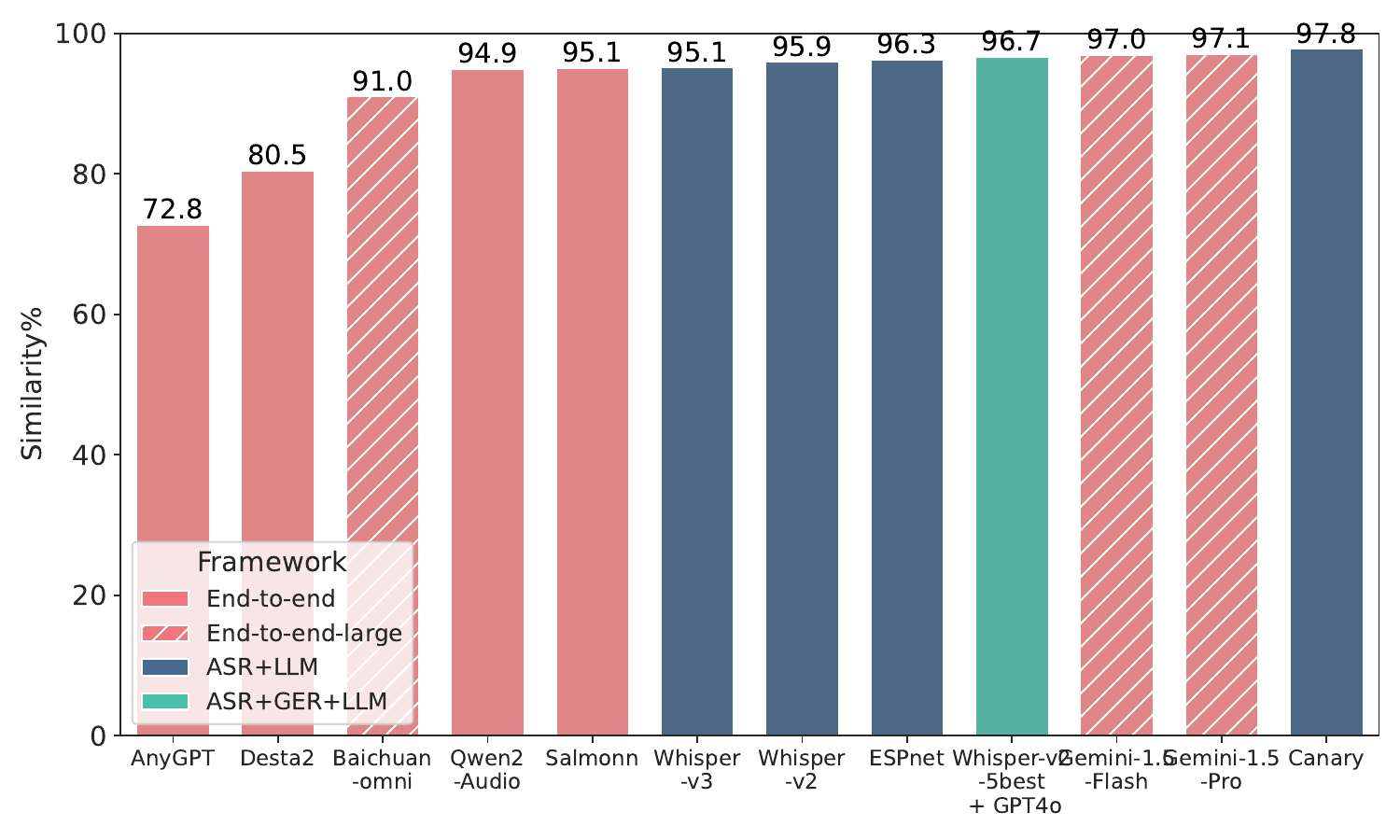}
  \caption{Similarity on voxpopuli}
  \label{sim_voxpopuli}
\end{figure}

As in Figure~\ref{sim_earning22}~\ref{sim_medasr}~\ref{sim_voxpopuli}, two architectures become competitive on each dataset, especially for end-to-end-large LLM$_\text{Voice}$s show strongest performance. Meanwhile, GER transfers to provice positive influence on ASR results by capturing the semantic meaning of the voice inputs.

\subsection{QA Accuracy Results}
\begin{figure}
  \includegraphics[width=\linewidth]{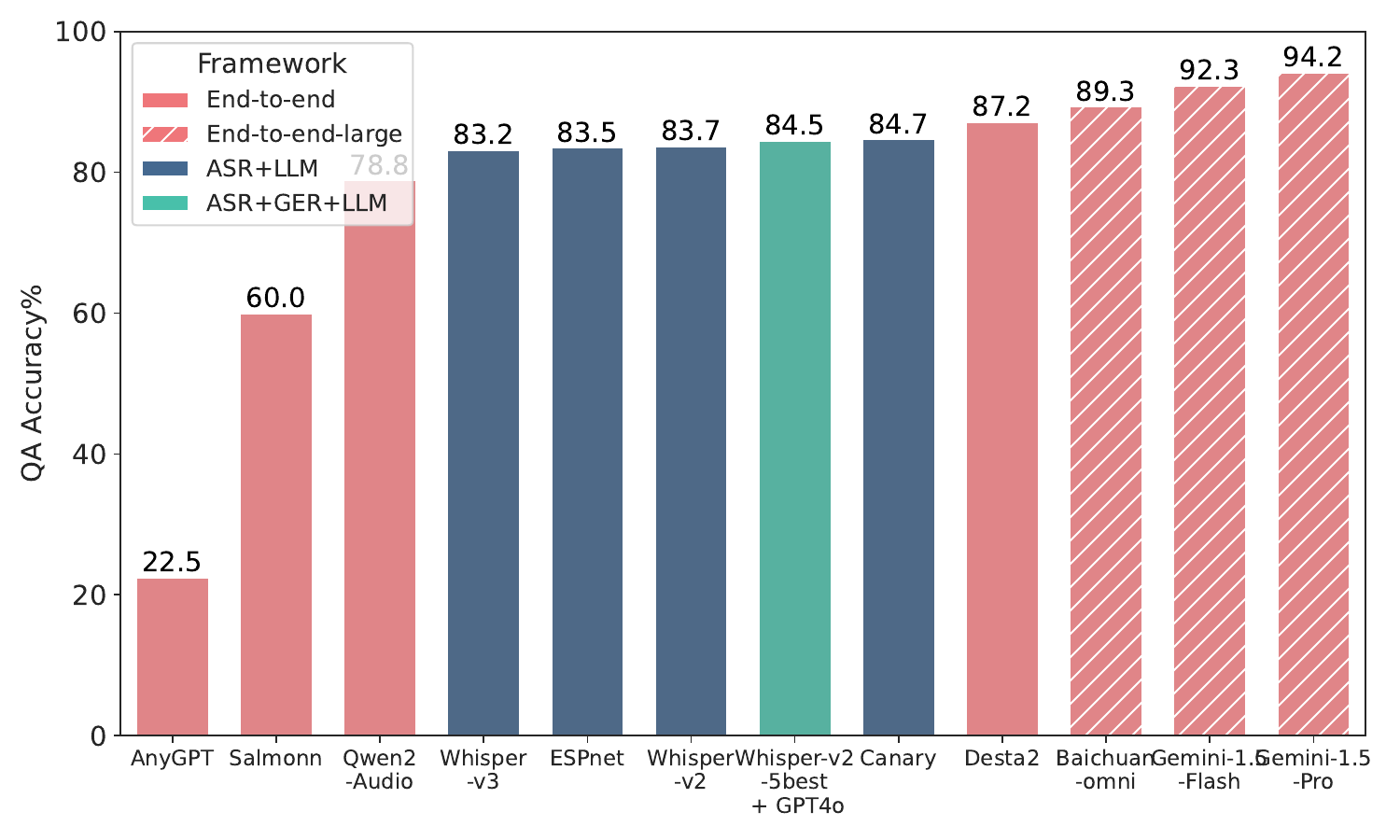}
  \caption{QA accuracy on earning22}
  \label{qa_earning22}
\end{figure}

\begin{figure}
  \includegraphics[width=\linewidth]{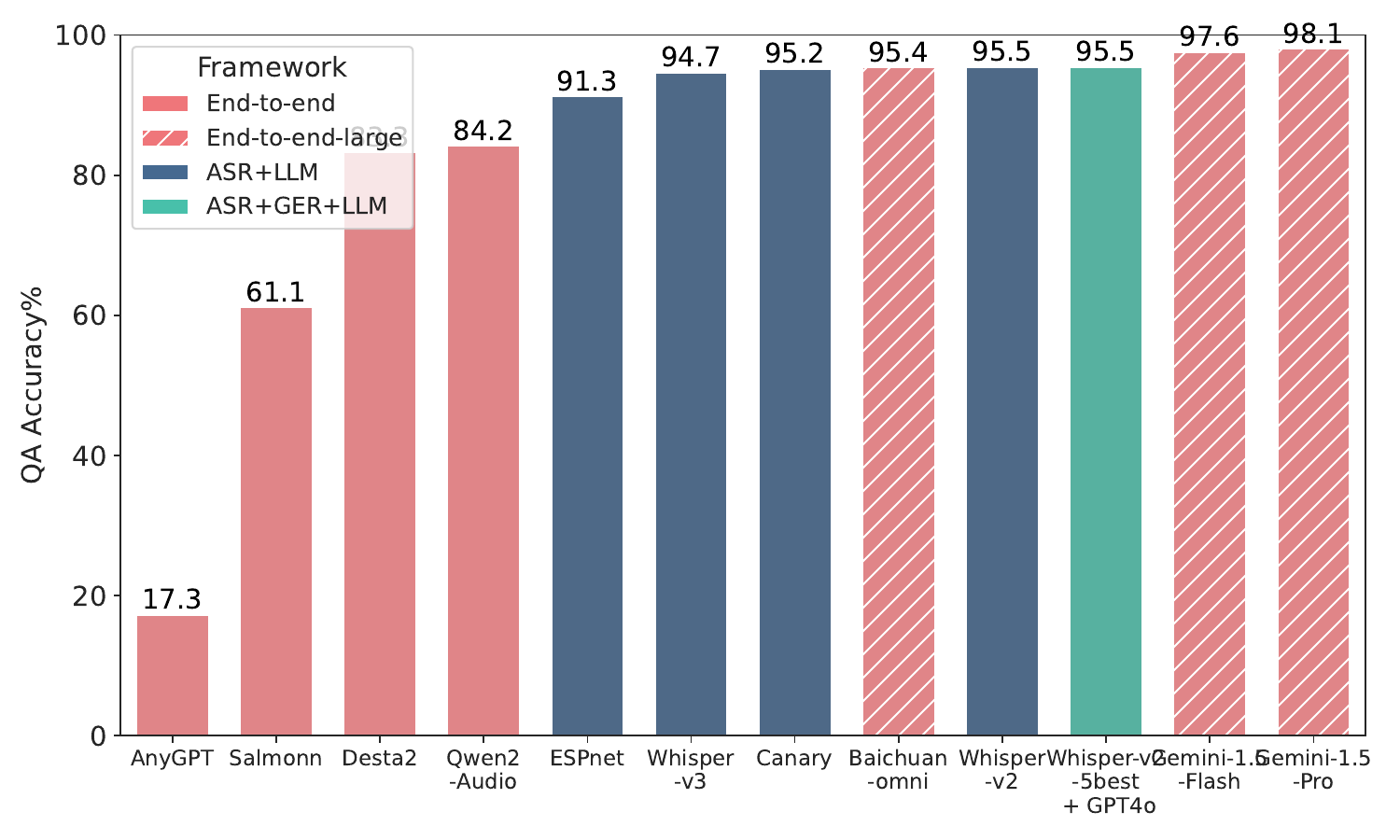}
  \caption{QA accuracy on medasr}
  \label{qa_medasr}
\end{figure}

\begin{figure}
  \includegraphics[width=\linewidth]{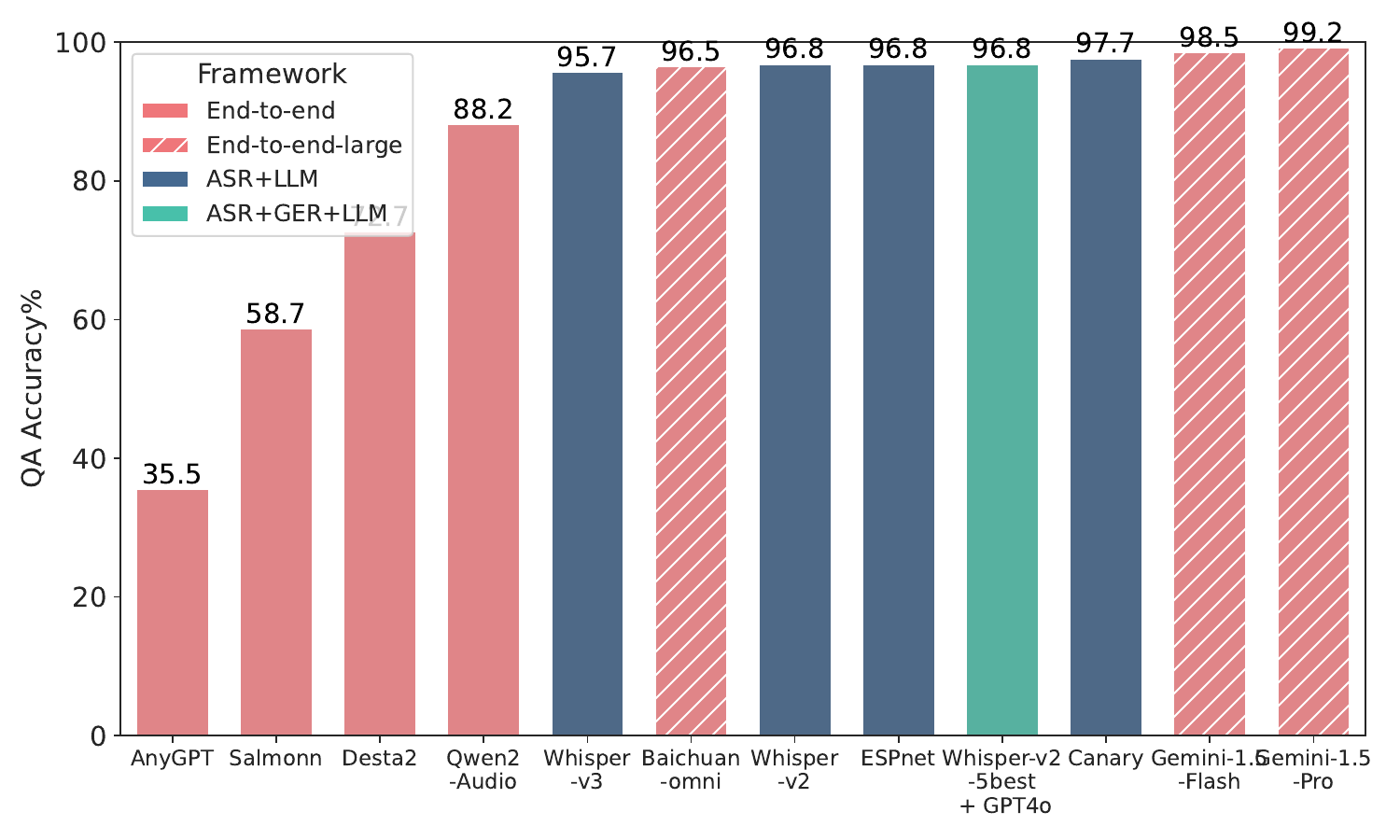}
  \caption{QA accuracy on voxpopuli}
  \label{qa_voxpopuli}
\end{figure}

As in Figure~\ref{qa_earning22}~\ref{qa_medasr}~\ref{qa_voxpopuli}, a interesting tendency occurs that end-to-end-large LLM$_\text{Voice}$s show the best performances in solving multi-choice questions while smaller end-to-end LLM$_\text{Voice}$s show the worst performances, with cascaded LLM$_\text{Voice}$s lying on the mid. This may indicate that modality alignment need scaling-up data size to train robust speech-text models and better than their cascaded foundation variants.

\section{Scaling Law Normalization}
\label{size normalization}
We realize that normalize the SIQ based on computations in training related to both model size and data size will be a significant step for fairly comparing various models. This normalization matches the age-aware normalization in computing human IQ, since models trained with less computations should be rewarded more SIQ if perform similar or better than larger models, representing the better training strategies in modality alignment. Thus, in our future work we will make efforts to include the scaling law normalization, even considering the fact that the computation cost is sometimes closed to acquire. 

\end{document}